%% file: paper.tex
\documentclass{article}

\usepackage{microtype}
\usepackage{graphicx}
\usepackage{subcaption}
\usepackage{booktabs} 
\usepackage{wrapfig}

\usepackage{hyperref}

\usepackage[accepted]{icml2026}

\usepackage{amsmath}
\usepackage{amssymb}
\usepackage{mathtools}
\usepackage{amsthm}
\usepackage[normalem]{ulem}
\usepackage[table]{xcolor}

\usepackage[capitalize,noabbrev]{cleveref}

\theoremstyle{plain}

\theoremstyle{definition}

\theoremstyle{remark}

\usepackage[textsize=tiny]{todonotes}
\usepackage{multirow}
\usepackage{makecell}

\icmltitlerunning{Probing Cross-modal Information Hubs in Audio-Visual LLMs}

\begin{document}
\input{preamble}

\twocolumn[
  \icmltitle{Probing Cross-modal Information Hubs in Audio-Visual LLMs}



  \icmlsetsymbol{equal}{*}

  \begin{icmlauthorlist}
    \icmlauthor{Jihoo Jung}{yyy}
    \icmlauthor{Chaeyoung Jung}{yyy}
    \icmlauthor{Ji-Hoon Kim}{xxx}
    \icmlauthor{Joon Son Chung}{yyy}
  \end{icmlauthorlist}

  \icmlaffiliation{yyy}{Department of Electrical Engineering, Korea Advanced Institute of Science and Technology (KAIST), Daejeon, Republic of Korea}
  \icmlaffiliation{xxx}{The Graduate School of Advanced Imaging Science, Multimedia \& Film, Chung-Ang University, Seoul, Republic of Korea}

  \icmlcorrespondingauthor{Joon Son Chung}{joonson@kaist.ac.kr}

  \icmlkeywords{Machine Learning, ICML}

  \vskip 0.3in
]



\printAffiliationsAndNotice{}  
\input{sec/abstract}
\input{sec/0_introduction}
\input{sec/1_related_works}
\input{sec/2_locating_token}
\input{sec/3_object_sink}
\input{sec/4_cross_modal_sink}

\input{sec/5_application}
\input{sec/6_conclusion}
\section*{Acknowledgement}
This work was supported by Institute of Information \& communications Technology Planning \& Evaluation (IITP) grant funded by the Korea government (MSIT) (RS-2025-
02215122, Development and Demonstration of Lightweight AI Model for Smart Homes).

\section*{Impact Statement}
This paper presents work whose goal is to advance the field
of Machine Learning. There are many potential societal
consequences of our work, none of which we feel must be
specifically highlighted here.
\nocite{langley00}

\bibliographystyle{icml2026}
\bibliography{shortstrings,refs}

\input{sec/appendix}

\end{document}

%% file: preamble.tex
\newcommand{\ra}{\textcolor[rgb]{0.00392156862745098, 0.45098039215686275, 0.6980392156862745}{\bf Hynj}}
\newcommand{\rb}{\textcolor[rgb]{0.00784313725490196, 0.6196078431372549, 0.45098039215686275}{\bf diPM}}
\newcommand{\rc}{\textcolor[rgb]{0.8352941176470589, 0.3686274509803922, 0.0}{\bf QfFh}}
\definecolor{uptri}{RGB}{38,114,168}
\definecolor{dntri}{RGB}{178,34,52}
\newcommand{\up}{\textcolor{uptri}{\ding{115}}}
\newcommand{\dn}{\textcolor{dntri}{\ding{116}}}
\definecolor{lightgray}{rgb}{0.83, 0.83, 0.83}
\definecolor{Gray}{gray}{0.6}
\definecolor{aliceblue}{rgb}{0.94, 0.97, 1.0}
\definecolor{mistyrose}{rgb}{1.0, 0.89, 0.88}
\definecolor{backcolour}{rgb}{0.95,0.95,0.92}

\newcommand{\newpara}[1]{\vspace{-1pt}\noindent\textbf{#1}}

\crefname{equation}{Eq.}{Eqs.}
\Crefname{equation}{Equation}{Equations}

\crefname{figure}{Fig.}{Figs.}
\Crefname{figure}{Figure}{Figures}

\crefname{table}{Tab.}{Tabs.}
\Crefname{table}{Table}{Tables}

\crefname{section}{Sec.}{Secs.}
\Crefname{section}{Section}{Sections}

\crefname{algorithm}{Alg.}{Algs.}
\Crefname{algorithm}{Algorithm}{Algorithms}

%% file: sec/abstract.tex
\begin{abstract}
Audio-visual large language models (AVLLMs) have recently emerged as a powerful architecture capable of jointly reasoning over audio, visual, and textual modalities. In AVLLMs, the bidirectional interaction between audio and video modalities introduces intricate processing dynamics, necessitating a deeper understanding of their internal mechanisms. However, unlike extensively studied text-only or large vision language models, the internal workings of AVLLMs remain largely unexplored. In this paper, we focus on cross-modal information flow between audio and visual modalities in AVLLMs, investigating where information derived from one modality is encoded within the token representations of the other modality. Through an analysis of multiple recent AVLLMs, we uncover two common findings. First, AVLLMs primarily encode integrated audio-visual information in sink tokens. Second, sink tokens do not uniformly hold cross-modal information. Instead, a distinct subset of sink tokens, which we term cross-modal sink tokens, specializes in storing such information. 
Based on these findings, we further propose a simple training-free  hallucination mitigation method by encouraging reliance on integrated cross-modal information within cross-modal sink tokens. Our code is available at \url{https://github.com/kaistmm/crossmodal-hub}.
\end{abstract}

%% file: sec/0_introduction.tex
\section{Introduction}
Recent advancements in large language models (LLMs) have catalyzed the rapid evolution of multimodal LLMs (MLLMs), extending text-centric capabilities to encompass diverse modalities~\citep{yu2024rlhf,weng2024longvlm, huang2024audiogpt}. Among these, audio-visual LLMs (AVLLMs), which integrate auditory and visual inputs via an LLM decoder to generate textual responses, are crucial for achieving a holistic understanding of real-world environments. By jointly processing audio-visual sensory streams, AVLLMs enable reasoning capabilities that closely mirror human multimodal perception, encompassing a comprehensive scope of multimodal context~\citep{zhang2023video, cheng2024videollama2advancingspatialtemporal}. 
As these models continue to evolve and gain widespread adoption, investigating their internal mechanisms has become an imperative step to ensure safety and robustness.

Extensive research has investigated the internal mechanisms of LLMs and LVLMs. Specifically, causal tracing has been utilized to track information transfer~\citep{basu2024understandinginformationstoragetransfer,kiciman2024causal, meng2023locatingeditingfactualassociations,li2025causaltracingobjectrepresentations}. In parallel, recent works have employed sparse autoencoders to interpret their internal representations~\citep{huben2024sparse, shi2025route,pach2025sparse, lim2025sparse}.
However, the internal dynamics of AVLLMs remain largely unexplored. This gap is particularly significant as the integration of the audio modality introduces a unique complexity, necessitating a deeper understanding of simultaneous audio-visual interaction. Elucidating these mechanisms is critical to not only advance model architectures but also verify the factual integrity of models in complex multimodal contexts.

\begin{figure}[t]
    \centering
    \vspace{-4mm}
    \includegraphics[width=0.9\linewidth]{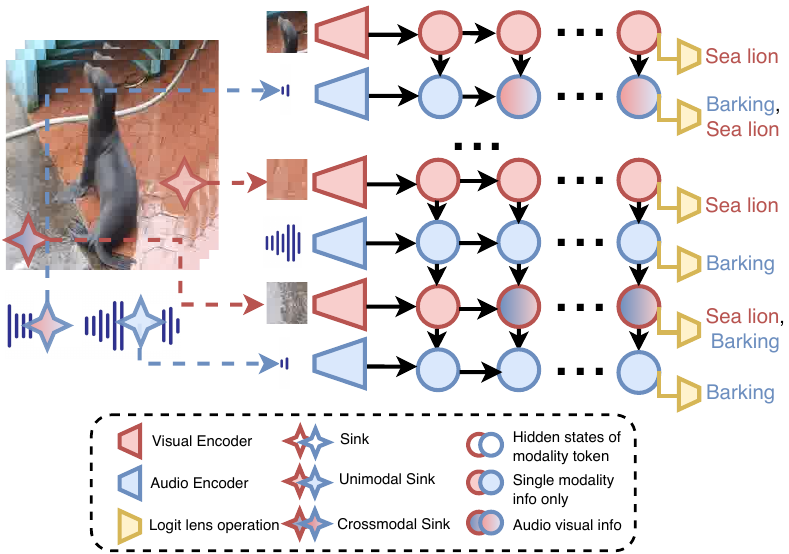}
    \vspace{-1mm}
    \caption{\textbf{Cross-modal information is primarily stored in cross-modal sink tokens.} Consider an audiovisual clip of a barking sea lion. Cross-modal sink tokens aggregate cues from both modalities, whereas unimodal sink tokens encode information solely from their native modality.}
    \label{fig:main}
    \vspace{-4mm}
   \vspace{-3mm}

\end{figure}
In this work, we explore the internal mechanisms of cross-modal interaction, specifically investigating where information derived from one modality (audio or visual) is stored within the token representations of the other modality in AVLLMs. To this end, we adapt the well-established causal tracing technique~\cite{meng2023locatingeditingfactualassociations, basu2024understandinginformationstoragetransfer} via our proposed unimodal dominance based framework. 
By analyzing instances where the model's output is governed by a single dominant modality, we can pinpoint specific tokens within the non-dominant modality that encapsulate information from the dominant modality.

Our key findings are twofold. First, we demonstrate that attention sink tokens-known to receive disproportionately high attention weights in LLMs and LVLMs-serve as the primary repositories for storing cross-modal information. Second, we find that these sink tokens do not uniformly hold such information. By distinguishing between unimodal sink tokens, which attract substantial attention from their own modality and cross-modal sink tokens, which are heavily attended to by the other modality, we demonstrate that cross-modal sink tokens serve as the distinct carriers of cross-modal information.
As illustrated in \cref{fig:main}, cross-modal sink tokens, characterized by blended colors, encapsulate comprehensive audio-visual concepts (e.g., ``barking" and ``sea lion"), whereas unimodal sink tokens capture information solely from their respective modality.

Finally, leveraging these mechanistic insights, we propose a simple, training-free method to mitigate object hallucinations in AVLLMs. By steering attention toward cross-modal sink tokens, our approach enhances audio-visual integration and significantly reduces hallucinations.

In summary, our contributions are:
\begin{itemize}
    \item We uncover that cross-modal information in AVLLMs is not uniformly distributed but is localized within attention sink tokens.
    \item We introduce a functional categorization of sink tokens into unimodal and cross-modal types, revealing that cross-modal sink tokens serve as the critical hubs for integrating audio-visual information.
    \item We propose a simple, training-free method that mitigates object hallucinations by strategically amplifying the influence of cross-modal sink tokens.
\end{itemize}

%% file: sec/1_related_works.tex
\section{Related Works}
\textbf{Audio-visual large language models.} Extending the capabilities of LLMs, AVLLMs have recently emerged, broadening text-centric understanding to encompass audio and visual perception~\cite{zhang2023video, cheng2024videollama2advancingspatialtemporal,chowdhury2024meerkat,lyu2023macaw,ye2024cat,tang2025video,guo2025aligned,xu2025qwen25omnitechnicalreport,xu2025qwen3omnitechnicalreport}.
By jointly processing visual and auditory signals, AVLLMs enable context-aware multimodal reasoning for complex real-world scenarios. Early works such as VideoLLaMA~\cite{zhang2023video} integrated audio-visual inputs into LLaMA~\cite{touvron2023llama}. Building on this, video-SALMONN~\cite{sunvideo} enhanced fine-grained temporal processing to improve speech understanding. More recently, Qwen2.5-Omni and Qwen3-Omni~\cite{xu2025qwen25omnitechnicalreport, xu2025qwen3omnitechnicalreport} advanced the field by introducing the capability to generate both text and natural speech as outputs. To effectively fuse audio, visual, and text modalities, various architectural approaches have been explored. Recent state-of-the-art AVLLMs~\cite{xu2025qwen25omnitechnicalreport,xu2025qwen3omnitechnicalreport,tang2025video,sunvideo_o1,guo2025aligned} process audio and visual inputs via distinct encoders, where the resulting audio and visual embeddings are temporally interleaved and concatenated with text embeddings to serve as input for the LLM backbone. We focus our analysis on this established architecture, aiming to investigate the underlying mechanisms of information exchange between audio and visual modalities. 

\textbf{Mechanistic interpretability.} Mechanistic interpretability aims to reverse engineer how neural networks process information and make decisions, with the goal of improving transparency, reliability, and trustworthiness. 
While recent works have leveraged various techniques to probe internal mechanisms, such as causal tracing~\cite{meng2023locatingeditingfactualassociations,kiciman2024causal,basu2024understandinginformationstoragetransfer,li2025causaltracingobjectrepresentations}, circuit discovery~\cite{wang2022interpretability,nandaprogress,he2025towards_circuit}, and sparse autoencoders~\cite{huben2024sparse,shi2025route,pach2025sparse,lim2025sparse} in LLMs and LVLMs, the inner workings of AVLLMs remain largely unexplored.
Unlike unimodal or bimodal LLMs, AVLLMs facilitate bidirectional information exchange between audio and visual modalities at the token level. These complex interactions introduce new dynamics, necessitating specialized interpretability analyses. 

\textbf{Causal tracing.} Among interpretability methods, causal tracing is a widely adopted technique that identifies the causal contribution of specific model components to the final predictions, grounded in causal mediation analysis~\cite{pearl2001direct,vig2020causal}. Specifically, the process entails three distinct forward passes: (i) a \emph{clean run} with the original input, (ii) a \emph{corrupted run} with a perturbed input designed to degrade the model’s prediction, and (iii) a \emph{corrupted-with-restoration run}, in which a hidden state from the clean run is patched into the corresponding position of the corrupted run. By quantifying the recovery of the clean prediction, this step determines whether the patched activation encodes critical information, enabling us to trace the causal flow.

%% file: sec/2_locating_token.tex
\section{Unimodal Dominance Framework for Tracing Cross-modal Information}
\label{sec:causal_tracing}
To trace bidirectional cross-modal information flow, we first introduce a unimodal dominance framework (\cref{sec:av_conflict}), followed by the integration of causal tracing into this setting (\cref{causal_tracing_frame}). Subsequently, we define a metric to quantify the influence of causal states (\cref{measure}).

\subsection{Unimodal Dominance Framework}
\label{sec:av_conflict} 
The unimodal dominance framework captures scenarios where a single modality governs the model's output by providing decisive cues, while its counterpart remains ambiguous. For instance, in identifying a tennis match, the visual modality governs the output by revealing distinct visual features such as racket, whereas the audio modality remains ambiguous as the impact sounds often resemble those of other sports. Positing that information from the dominant modality is propagated into the non-dominant stream via self-attention within transformer block, we aim to localize these propagated signals within the non-dominant stream. Specifically, we leverage audio-dominant samples to identify which visual tokens primarily encode audio-derived information, and video-dominant samples to determine which audio tokens encode video-derived cues.

We classify an instance as \emph{Audio-Dominant} when the joint prediction $\hat{y}_{av}$ aligns with the audio-only prediction $\hat{y}_{a}$ but diverges from the erroneous video-only prediction $\hat{y}_{v}$:

\begin{equation}
\hat{y}_{av} = \hat{y}_a \neq \hat{y}_v.
\label{eq:audio_dominant}
\end{equation}
Conversely, the \emph{Video-Dominant} setting is identified when the joint prediction aligns with the video-only prediction but differs from the audio counterpart:
\begin{equation}
\hat{y}_{av} = \hat{y}_v \neq \hat{y}_a.
\label{eq:video_dominant}
\end{equation}

\subsection{Causal Tracing under Unimodal Dominance Framework}
\label{causal_tracing_frame}
\begin{figure*}[t]
    \centering
\includegraphics[width=0.9\linewidth]{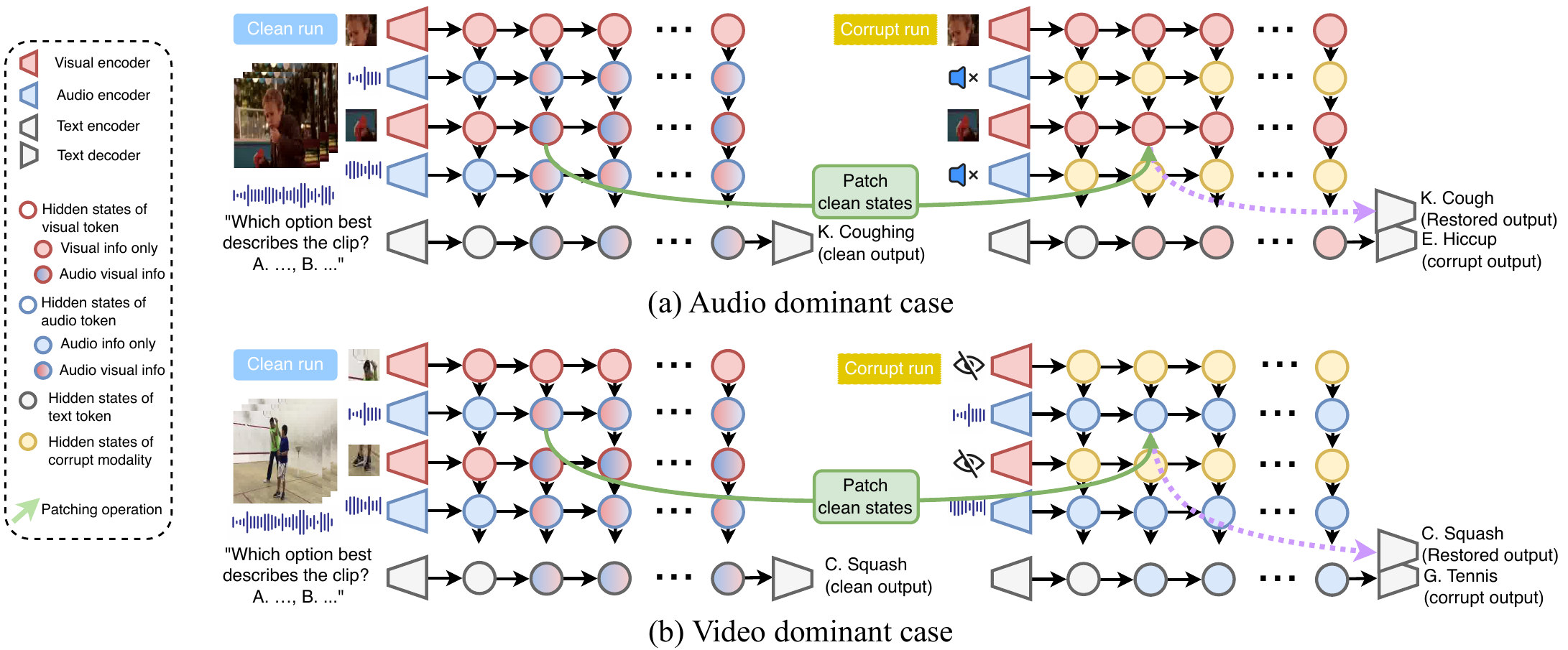}
\vspace{-1mm}
   \caption{\textbf{Causal Tracing under the Unimodal Dominance Framework.}
For the audio-dominant case, the corrupt run is constructed by corrupting the audio modality, and restoration is conducted by patching hidden states of video tokens from the clean run. Conversely, for the video-dominant case, the video modality is corrupted, and hidden states of audio tokens are patched from the clean run. We expect patching the non-dominant tokens to partially recover the clean prediction, as these tokens encode information transferred from the dominant modality.}
    \label{fig:causal_tracing_figure}
    \vspace{-4mm}
\end{figure*}
\cref{fig:causal_tracing_figure} illustrates causal tracing under unimodal dominance framework with examples.
In the \emph{clean run}, the model receives both audio and video inputs without modification, allowing full cross-modal interaction. The model is prompted with a multiple-choice question (e.g., “Which option best describes the clip? A. People farting B. People humming...”) and yields the clean output $o_{\text{clean}}$.

To construct the corrupted run, we perturb the dominant modality’s input tokens by zeroing out their raw representations before they are fed into the corresponding modality encoder. As a result, the model is forced to rely solely on the non-dominant modality token set, $\mathtt{Tokens}_{\text{nondom}}$, which encodes misleading semantic information, thereby producing an incorrect prediction $o_{\text{corrupt}}$.

We construct a \emph{corrupted-with-restoration run} by patching the hidden states from the clean run ($h_{S}^{\text{clean}}$) into the corrupted run for a subset of tokens $S \subset \mathtt{Tokens}_{\text{nondom}}$. This patching is applied to the hidden states feeding into the self-attention sublayer in every Transformer block. Details on the selection of patching locations are provided in Appendix~\ref{appendix:causal_tracing_further}. Patching $h_{S}^{\text{clean}}$ is expected to partially recover the clean prediction, as $h_{S}^{\text{clean}}$ encodes information originating from the dominant modality via cross-modal interactions.

\subsection{Metrics for Causal Tracing}
\label{measure}
We quantify the amount of cross-modal information encoded in the hidden states of the token subset $S$ by measuring changes in output probabilities between the corrupted run and the restored run.
Let $P[o]$ and $P_{h_S^{\mathrm{clean}}}[o]$ denote the probability of generating output $o$ under the corrupted run and the restored run, respectively. Following~\cite{meng2023locatingeditingfactualassociations}, we define the \emph{indirect effect} (IE) with respect to the clean output $o_{\text{clean}}$ as
\begin{equation}
\mathrm{IE}_{\text{clean}}(S)
=
P_{h_S^{\mathrm{clean}}}\!\left[o_{\text{clean}}\right]
-
P\!\left[o_{\text{clean}}\right].
\label{eq:ie_clean}
\end{equation}
A high $\mathrm{IE}_{\text{clean}}(S)$ indicates that $S$ encodes a great amount of information originating from the dominant modality.

We also define the indirect effect with respect to the corrupted output $o_{\text{corrupt}}$ as
\begin{equation}
\mathrm{IE}_{\text{corrupt}}(S)
=
P\!\left[o_{\text{corrupt}}\right]
-
P_{h_S^{\mathrm{clean}}}\!\left[o_{\text{corrupt}}\right].
\label{eq:ie_corrupt}
\end{equation}
A high $\mathrm{IE}_{\text{corrupt}}(S)$ indicates that $S$  incorporates significant signals from the dominant modality, thereby effectively overriding the ambiguous cues of the non-dominant modality.

Consequently, we utilize these metrics to identify which tokens serve as key mediators of cross-modal information flow; higher values indicate that the subset of token $S$ acts as a critical repository of cross-modal information.

%% file: sec/3_object_sink.tex
\section{Where Is Cross-modal Information Located?}
Using the causal tracing framework introduced in \cref{sec:causal_tracing}, we now investigate which tokens primarily encode cross-modal information in AVLLMs.

\subsection{Hypothesis}
Drawing upon recent findings in LVLMs on information storage, we formulate two competing hypotheses: object-centric localization and sink-centric localization

\textbf{Object-centric localization.}
\cite{neotowards} demonstrate that object-centric information is primarily stored in object tokens—tokens corresponding to spatial locations of the object in the original image in LVLMs. This suggests that cross-modal information might be localized within object-aligned tokens, arising from the interaction between the object representation in one modality and the corresponding object in the complementary modality. Here, we define \emph{audio object tokens} as those capturing the temporal segments of the object's sound, and \emph{video object tokens} as those encoding its spatio-temporal visual regions.

\textbf{Sink-centric localization.} In contrast, \cite{darcetvision, luo2025sinksinkvisualinformation} demonstrate that attention sink aggregate high-level global visual information in LVLMs. This suggests that cross-modal information might be localized within sink tokens, acting as the primary locus for abstract semantic cues derived from both modalities.
This hypothesis fundamentally contrasts with the object-centric view, as sink tokens are known to emerge from non-object, background positions in LVLMs.
For clarity, we refer to sink tokens in the audio modality as \emph{audio sink tokens} and those in the video modality as \emph{video sink tokens}.

\subsection{Analysis Configuration}
\label{exp}
To validate these hypotheses, we perform causal tracing experiments across three recent open-source AVLLMs: Qwen2.5-Omni (7B/3B)~\cite{xu2025qwen25omnitechnicalreport}, video-SALMONN-o1 (7B)~\cite{sunvideo_o1}, and video-SALMONN2+ (7B/3B)~\cite{tang2025video}.             

\textbf{Dataset.}
We use a subset of the VGGSound test set~\cite{chen2020vggsoundlargescaleaudiovisualdataset} to construct audio-dominant and video-dominant evaluation cases.
VGGSound is a large-scale audio–visual dataset consisting of in-the-wild videos with class label. 
Building on findings that certain categories exhibit distinct modality dominance~\cite{jiang2025bridgingearseyesanalyzing}, we select two disjoint sets of 20 classes, each with 1,000 videos, 
as candidate pools for the audio-dominant and video-dominant settings, respectively. For each sample, we present 20 candidate options and ask the model to select the option that best describes the clip. We then retain only the samples that satisfy Eq.~\eqref{eq:audio_dominant} and Eq.~\eqref{eq:video_dominant} for the corresponding dominance setting, and use the filtered set as the final evaluation data for each model. Further details are provided in Appendix~\ref{appendix:dataset_for_causal}.

\textbf{Token identification.} We employ a recent image segmentation model~\cite{ravisam} and a sound event detection model~\cite{wuflam} to identify video and audio object tokens, respectively.
For sink tokens, we adopt the definition from LVLMs~\cite{kang2025toldvisualattentionsink, luo2025sinksinkvisualinformation},
where sinks are characterized by abnormally large activations in predefined sink dimensions, with one crucial modification.
Prior works determine sink tokens on a per-layer basis and report that sink locations are stable across layers.
In contrast, we observe substantial layer-wise variation of sink tokens in AVLLMs (Appendix~\ref{app:definition_of_sink_tokens}).
 Consequently, we define global sink tokens as the top-$\frac{|\mathcal{T}|}{N}$ tokens with the highest occurrence frequency across all layers, where $|\mathcal{T}|$ and $N$ denote the input sequence length and selection sparsity, respectively.

\textbf{Patching strategies.}
We measure $\mathrm{IE}_{\text{clean}}$ and $\mathrm{IE}_{\text{corrupt}}$ across four distinct patching scenarios:
(1) \emph{All non-dominant modality tokens} ($\mathtt{Tokens}_{\text{nondom}}$), which serves as an upper bound on performance recovery;
(2) \emph{Object tokens}, using video object tokens for audio-dominant cases and audio object tokens for video-dominant cases;
(3) \emph{Sink tokens}, using video sink tokens for audio-dominant cases and audio sink tokens for video-dominant cases.
We report results for $N \in \{2,3,4\}$ to control the number of sink tokens;
and (4) \emph{Random tokens}, which serve as a baseline and are sampled to match the number of sink tokens.

\subsection{Finding1: Cross-modal Information is Primarily Encoded in Sink Tokens}
\input{table/3.3_result} 
\cref{tab:object_sink} summarizes the results.
Across all models in audio-dominant settings, restoring sink tokens consistently yields larger $\mathrm{IE}_{\text{clean}}$ and $\mathrm{IE}_{\text{corrupt}}$ than restoring object tokens or randomly selected tokens with a comparable number of tokens. 
Notably, sink tokens maintain substantially higher IE values than random tokens even when the number of sink tokens is small ($N=3$ or $N=4$). 
In video-dominant settings, although the effect is less pronounced for video-SALMONN2+ due to the small number of patched audio tokens, we observe the same overall trend. 
\uline{These results indicate that, in both directions of information storage, cross-modal information is primarily encoded in sink tokens rather than in object tokens or being uniformly distributed across tokens.}

%% file: table/3.3_result.tex
\begin{table*}[t]
\centering
\tiny
\fontsize{5.5pt}{6.5pt}\selectfont
\caption{\textbf{Patching results across distinct token sets}: all non-dominant modality tokens, object tokens, sink tokens ($N=2,3,4$), and random tokens. We report $\mathrm{IE}_{\text{clean}}$, $\mathrm{IE}_{\text{corrupt}}$, and the number of patched tokens to ensure a fair comparison. \textbf{Bold} and \underline{underlined} values indicate the best and second-best results, respectively; note that the ``all non-dominant tokens'' case is excluded from this ranking as it serves as an empirical upper bound.}
\vspace{-1mm}

\label{tab:object_sink}
\scriptsize
\setlength{\tabcolsep}{3pt}

\newcolumntype{C}[1]{>{\centering\arraybackslash}p{#1}}
\resizebox{\textwidth}{!}
{
    \begin{tabular}{ll|
    C{0.78cm}C{0.75cm}>{\columncolor{gray!15}}C{0.65cm}|   
    C{0.78cm}C{0.75cm}>{\columncolor{gray!15}}C{0.65cm}|   
    C{0.78cm}C{0.75cm}>{\columncolor{gray!15}}C{0.65cm}|   
    C{0.78cm}C{0.75cm}>{\columncolor{gray!15}}C{0.65cm}|   
    C{0.78cm}C{0.75cm}>{\columncolor{gray!15}}C{0.65cm}    
    }
    \toprule
    \textbf{Modality} & \textbf{Ablation}
    & \multicolumn{3}{c|}{\textbf{Qwen2.5-Omni(7B)}}
    & \multicolumn{3}{c|}{\textbf{Qwen2.5-Omni(3B)}}
    & \multicolumn{3}{c|}{\textbf{video-SALMONN-o1(7B)}}
    & \multicolumn{3}{c|}{\textbf{video-SALMONN2+(7B)}}
    & \multicolumn{3}{c}{\textbf{video-SALMONN2+(3B)}} \\
    
    \cmidrule(lr){3-5}
    \cmidrule(lr){6-8}
    \cmidrule(lr){9-11}
    \cmidrule(lr){12-14}
    \cmidrule(lr){15-17}

    &
    & \tiny IE$_{\text{clean}}$ $\uparrow$
    & \tiny IE$_{\text{corr}}$ $\uparrow$
    & \tiny \#Tokens
    & \tiny IE$_{\text{clean}}$ $\uparrow$
    & \tiny IE$_{\text{corr}}$ $\uparrow$
    & \tiny \#Tokens
    & \tiny IE$_{\text{clean}}$ $\uparrow$
    & \tiny IE$_{\text{corr}}$ $\uparrow$
    & \tiny \#Tokens
    & \tiny IE$_{\text{clean}}$ $\uparrow$
    & \tiny IE$_{\text{corr}}$ $\uparrow$
    & \tiny \#Tokens
    & \tiny IE$_{\text{clean}}$ $\uparrow$
    & \tiny IE$_{\text{corr}}$ $\uparrow$
    & \tiny \#Tokens \\
    \midrule
    
    \multirow{9}{*}{\makecell{\textbf{Audio}\\\textbf{Dominant}}}
    & All
    & 9.61 & 5.28 & 1440
    & 7.83 & 3.48 & 1440
    & 35.55	&	33.18	&	1820
    & 6.45 & 5.27 & 1210
    &   1.92   &  2.15    &   1210   \\
    \addlinespace[0.2em]
    \cmidrule[0.5pt](lr){2-17}
    
    & Object
    & 5.04 & 2.44 & 613
    & 3.53 & 1.12 & 580
    & 16.22	&	15.06	&	852
    & 3.78 & 3.93 & 500
    &  0.72    &  1.16    &    447  \\
    
    \cmidrule(lr){2-17}
    
    & Sink (N=2)
    & \textbf{6.24} & \textbf{2.94} & 603
    & \textbf{6.99} & \textbf{2.70} & 605
    & \textbf{25.33}	&	\textbf{22.73}	&	818
    & \textbf{4.79} & \textbf{4.20} & 565
    & \textbf{1.33}  &  \textbf{1.38}  & 506   \\
    
    & Sink (N=3)
    & \underline{4.31} & 1.94 & 362
    & \underline{6.36} & \underline{2.08} & 354
    & \underline{21.42}	&	\underline{19.67}	&	514
    & 3.73 & 3.49 & 360
    &  0.93    &   0.94   &   297   \\
    
    & Sink (N=4)
    & 3.26 & 1.23 & 256
    & 5.50 & 1.64 & 243
    & 19.10	&	17.79	&	364
    & 3.23 & 3.33 & 256
    &  0.69    &   0.65   &   195   \\
    \addlinespace[0.2em]
    \cmidrule(lr){2-17}
    
    & Random (N=2)
    & 4.24 & \underline{2.37} & 603
    & 4.05 & 1.20 & 605
    & 20.43	&	18.11	&	818
    & \underline{4.21} & \underline{4.01} & 565
    &  \underline{ 1.09}   &  \underline{ 0.95  } &  506    \\
    
    & Random (N=3)
    & 2.97 & 1.55 & 362
    & 2.71 & 0.72 & 354
    & 12.65	&	13.54	&	514
    & 3.12 & 3.51 & 360
    &  0.67    &   0.74   &  297    \\
    
    & Random (N=4)
    & 1.93 & 0.87 & 256
    & 1.87 & 0.65 & 243
    &8.70	&	8.77	&	364
    & 3.02 & 3.33 & 256
    &  0.66    &  0.53    &   195   \\
    
    \midrule[0.8pt]
    
    \multirow{9}{*}{\makecell{\textbf{Video}\\\textbf{Dominant}}}
    & All
    & 8.21 & 13.63 & 249
    & 2.43 & 8.85  & 249
    & 3.63	&	4.08	&	153
    &  0.46 & 1.86 & 60    
    &   -0.05   &   -0.04    &   60   \\
    \addlinespace[0.2em]
    \cmidrule[0.5pt](lr){2-17}
    
    & Object
    &    \underline{4.97}	&	\underline{8.44	}&	149
    &    1.59	&	\underline{6.41}	&	149
    &    2.07	&	0.40	&	78
    &    \underline{0.22}&\underline{1.71}	&7
    &    \underline{-0.01}	&	-0.06	&	7    \\
    \addlinespace[0.2em]
    \cmidrule(lr){2-17}
    
    & Sink (N=2)
    & \textbf{5.47} & \textbf{8.54} & 144
    & \textbf{2.07} & \textbf{ 6.87} & 147
    & \textbf{3.57}	&	\textbf{3.87}	&	117
    &   \textbf{0.28} & \textbf{2.24} &   9   
    &   -0.02   &    0.00  &   29   \\
    
    & Sink (N=3)
    & 4.40 & 7.12 & 86
    & \underline{1.62} & 5.88 & 109
    & \underline{3.45}	&	\underline{3.66}	&	76
    &  0.08    & 1.70     & 3 
    &   -0.03   &   -0.06   &  16    \\
    
    & Sink (N=4)
    & 3.10 & 6.28 & 60
    & 1.10 & 4.78 & 85
    &3.30	&	3.28	&	52
    & 0.06     & 1.29  & 2
    &  \underline{-0.01 }   & \textbf{0.07}     &    13  \\
    \addlinespace[0.2em]
    \cmidrule(lr){2-17}
    
    & Random (N=2)
    & 4.56 & 6.83 & 144
    & 1.22 & 5.29 & 147
    & 2.86	&	2.22	&	117
    &  0.21    &   \underline{1.77}   & 9
    &  -0.02    & 0.01     &    29  \\
    
    & Random (N=3)
    & 2.70 & 3.62 & 86
    & 0.94 & 4.34 & 109
    &1.69	&	-0.44	&	76
    &   0.12   & 1.28     & 3
    &  \textbf{}{0.00}    &   \underline{0.02}     &    16  \\
    
    & Random (N=4)
    & 1.86 & 2.30 & 60
    & 0.58 & 3.46 & 85
    & 0.9	&	-2.52	&	52
    & 0.21     &    1.28  &   2  
    &  \underline{-0.01 }   & \underline{0.02}    & 13\\

    \bottomrule
    \end{tabular}
}
\vspace{-3mm}
\end{table*}

%% file: sec/4_cross_modal_sink.tex
\section{Are Sink Tokens Homogeneous Cross-modal Information Holders?}
Motivated by the finding that visual sink tokens aggregate information from the visual tokens attending to them in LVLMs~\cite{darcetvision,luo2025sinksinkvisualinformation}, we analyze the incoming attention to sink tokens. We specifically investigate the existence of modality-specific attention biases-whether individual sink tokens preferentially aggregate attention from their own modality or the complementary modality. Upon establishing the existence of these biases, we further investigate whether the subset of sink tokens that primarily attract attention from the complementary modality specializes in holding cross-modal information.
  
\subsection{Dissecting Sink Tokens} 
\label{sec:dissec}
To examine modality-specific attention biases, we first quantify how much incoming attention each sink token receives from video versus audio tokens. Concretely, we define the \emph{Modality Dominance Score} (MDS) for a sink token $i$ at layer $l$, which measures the relative dominance of video attention over audio attention:
\begin{equation}
\mathrm{MDS}_{i}^{l}
=
\frac{
\bar{A}_{\text{video}, i}^{l} - \bar{A}_{\text{audio}, i}^{l}
}{
\bar{A}_{\text{video}, i}^{l} + \bar{A}_{\text{audio}, i}^{l}
}.
\end{equation}
Here, $\bar{A}_{\text{video}, i}^{l}$ and $\bar{A}_{\text{audio}, i}^{l}$ denote the mean attention scores received by sink token $i$ from the video and audio modality tokens at layer $l$, respectively.

\begin{figure}[t]
        \centering
        \includegraphics[width=0.8\linewidth]{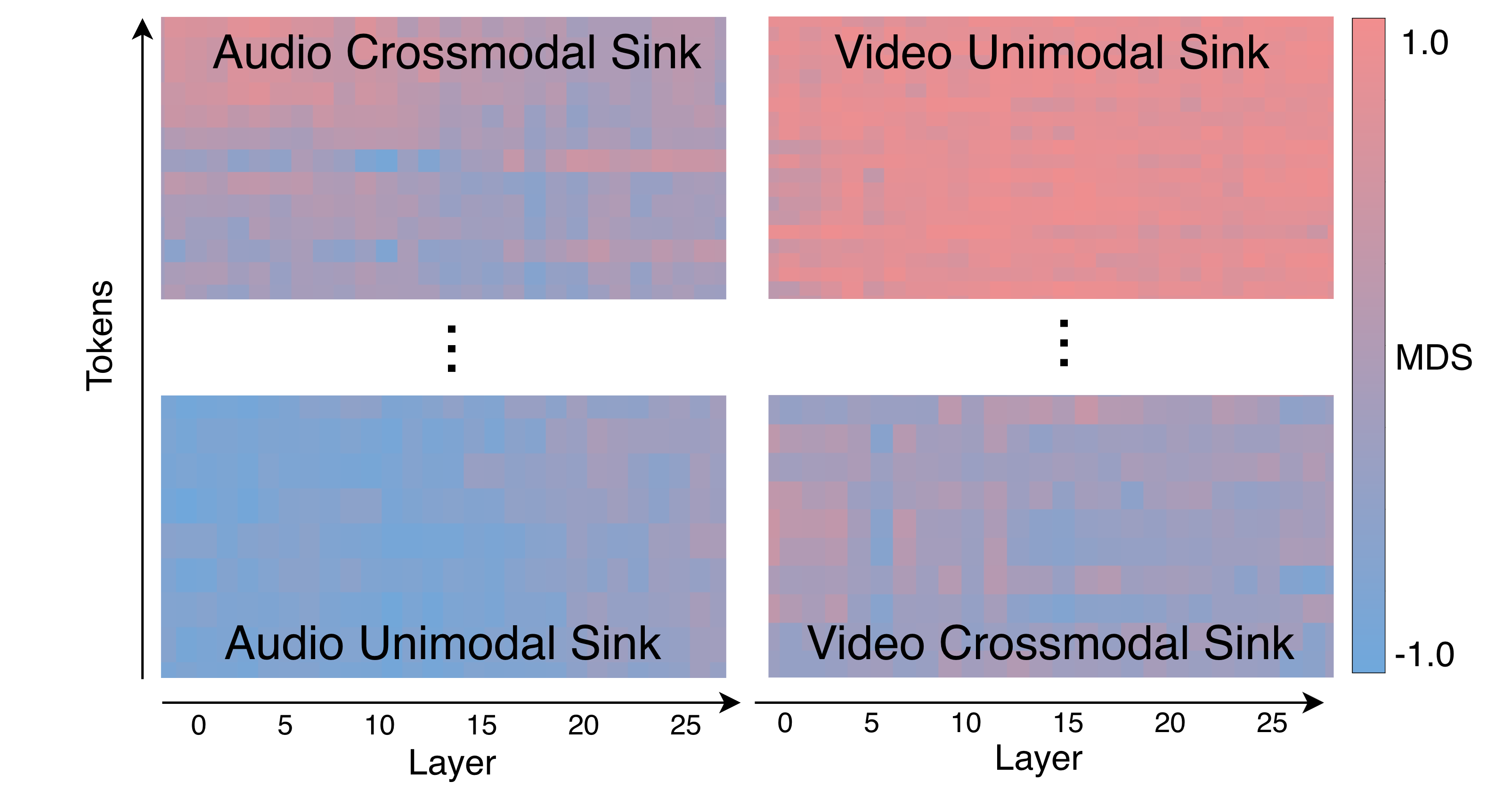}
        \vspace{-1mm}
\caption{\textbf{Layer-wise MDS of audio (left) and video (right) sink tokens, sorted by the layer-averaged MDS on Qwen2.5-Omni(7B).} Even within audio sink tokens, some tokens predominantly receive attention from the video modality (audio cross-modal sinks; high MDS), while others receive stronger attention from the audio modality (audio unimodal sinks; low MDS). A similar functional heterogeneity is observed for video sink tokens.}
\vspace{-7mm}
    \label{fig:mds_visualize}
\end{figure}

\cref{fig:mds_visualize} visualizes MDS values for audio and video sink tokens for a representative example. This visualization reveals that sink tokens may diverge into two groups: some receive incoming attention primarily from their own modality, while others from the other modality. See Appendix \ref{app:mds_stat} for further analysis on MDS.
    
Based on these observations, we investigate whether sink tokens receiving high attention from the complementary modality serve as specialized cross-modal information holders. For a fair comparison, we partition the tokens into two equal-sized groups: unimodal sink tokens, which primarily receive attention from their own modality, and cross-modal sink tokens, which are characterized by high attention from the complementary modality. We then replicate the causal tracing in \cref{exp} for each group to quantify their distinct contributions.

\subsection{Finding2: Cross-modal Sink Tokens Serve as the Primary Carriers of Cross-modal Information}
\input{table/4.2_result}
\cref{tab:unimodal_crossmodal} summarizes the results. Across models and modality dominant settings, restoring cross-modal sink tokens yields substantially high $\mathrm{IE}_\text{clean}$ and $\mathrm{IE}_\text{corrupt}$ scores, vastly outperforming their unimodal counterparts. Remarkably, results from cross-modal sink patching are comparable to those of the full sink set, implying that the observed causal effect of sink tokens is largely attributable to these cross-modal components. \uline{This trend indicates that sink tokens are heterogeneous in their information content: cross-modal sink tokens preferentially store and mediate cross-modal information, whereas unimodal sink tokens do not.} 
Additional analyses are provided in Appendix~\ref{appendix:causal_tracing_further}.

%% file: table/4.2_result.tex
\begin{table*}[t]
\centering
\caption{
\textbf{Patching results on unimodal and cross-modal token sets.} We additionally include sink token results for a comprehensive baseline.
We report $\mathrm{IE}_{\text{clean}}$, $\mathrm{IE}_{\text{corrupt}}$, and the number of patched tokens to ensure a fair comparison.
\textbf{Bold} and \underline{underlined} values indicate the best and second-best results, respectively, compared within each block of the same patch size ($N$) across sink, unimodal, and cross-modal tokens.
}
\vspace{-1mm}
\label{tab:unimodal_crossmodal}
\scriptsize
\setlength{\tabcolsep}{3pt}

\newcolumntype{C}[1]{>{\centering\arraybackslash}p{#1}}
\resizebox{\textwidth}{!}
{
    \begin{tabular}{ll|
    C{0.78cm}C{0.75cm}>{\columncolor{gray!15}}C{0.65cm}|   
    C{0.78cm}C{0.75cm}>{\columncolor{gray!15}}C{0.65cm}|   
    C{0.78cm}C{0.75cm}>{\columncolor{gray!15}}C{0.65cm}|   
    C{0.78cm}C{0.75cm}>{\columncolor{gray!15}}C{0.65cm}|   
    C{0.78cm}C{0.75cm}>{\columncolor{gray!15}}C{0.65cm}    
    }
    \toprule
    \textbf{Modality} & \textbf{Ablation}
    & \multicolumn{3}{c|}{\textbf{Qwen2.5-Omni(7B)}}
    & \multicolumn{3}{c|}{\textbf{Qwen2.5-Omni(3B)}}
    & \multicolumn{3}{c|}{\textbf{video-SALMONN-o1(7B)}}
    & \multicolumn{3}{c|}{\textbf{video-SALMONN2+(7B)}}
    & \multicolumn{3}{c}{\textbf{video-SALMONN2+(3B)}} \\
    
    \cmidrule(lr){3-5}
    \cmidrule(lr){6-8}
    \cmidrule(lr){9-11}
    \cmidrule(lr){12-14}
    \cmidrule(lr){15-17}

    &
    & \tiny IE$_{\text{clean}}$ $\uparrow$
    & \tiny IE$_{\text{corr}}$ $\uparrow$
    & \tiny \#Tokens
    & \tiny IE$_{\text{clean}}$ $\uparrow$
    & \tiny IE$_{\text{corr}}$ $\uparrow$
    & \tiny \#Tokens
    & \tiny IE$_{\text{clean}}$ $\uparrow$
    & \tiny IE$_{\text{corr}}$ $\uparrow$
    & \tiny \#Tokens
    & \tiny IE$_{\text{clean}}$ $\uparrow$
    & \tiny IE$_{\text{corr}}$ $\uparrow$
    & \tiny \#Tokens
    & \tiny IE$_{\text{clean}}$ $\uparrow$
    & \tiny IE$_{\text{corr}}$ $\uparrow$
    & \tiny \#Tokens \\
    
    \midrule

    \multirow{9}{*}{\makecell{\textbf{Audio}\\\textbf{Dominant}}}
    
    & Sink (N=2)
    & \textbf{6.24} & \textbf{2.94} & 603
    & \textbf{6.99} & \textbf{2.70} & 605
    & \textbf{25.33} & \textbf{22.73} & 818
    & \textbf{4.79} & \textbf{4.20} & 565
    & \textbf{1.33} & \textbf{1.38} & 506 \\
    
    & \quad Unimodal (N=2)
    & 0.65 & 0.23 & 301
    & 0.89 & 0.31 & 302
    & 7.25 & 6.82 & 409
    & 2.32 & 3.03 & 282
    & 0.21 & 0.45 & 252 \\
    
    & \quad Crossmodal (N=2)
    & \underline{5.58} & \underline{2.95} & 301
    & \underline{6.57} & \underline{2.33} & 302
    & \underline{21.30} & \underline{19.93} & 409
    & \underline{4.16} & \underline{3.69} & 282
    & \underline{1.27} & \underline{1.14} & 252 \\

    \addlinespace[0.15em]
    \cmidrule(lr){2-17}
    
    & Sink (N=3)
    & \textbf{4.31} & \textbf{1.94} & 362
    & \textbf{6.36} & \textbf{2.08} & 354
    & \textbf{21.42} & \textbf{19.67} & 514
    & \textbf{3.73} & \textbf{3.49} & 360
    & \textbf{0.93} & \textbf{0.94} & 297 \\

    &\quad  Unimodal (N=3)
    & 0.92 & 0.39 & 181
    & 1.02 & 0.18 & 177
    & 7.02	&	7.03	&	257
    & 2.06 & 2.87 & 180
    & 0.19   & 0.42   & 148 \\
    
    & \quad Crossmodal (N=3)
    & \underline{3.54} & \underline{1.52} & 181
    & \underline{5.73} & \underline{1.85} & 177
    & \underline{16.81} & \underline{15.78} & 257
    & \underline{3.35} & \underline{3.20} & 180
    & \underline{0.77} & \underline{0.70} & 148 \\

    \addlinespace[0.15em]
    \cmidrule(lr){2-17}
    
    & Sink (N=4)
    & \textbf{3.26} & \textbf{1.23} & 256
    & \textbf{5.50} & \textbf{1.64} & 243
    & \textbf{19.10}	& \textbf{17.79}	& 364
    & \textbf{3.23} & \textbf{3.33} & 256
    & \underline{0.69}  & \textbf{0.65}  & 195\\
    
    &\quad Unimodal (N=4)
    & 0.71 & 0.36 & 128
    & 1.07 & 0.32 & 121
    & 6.19 & 6.10 & 182
    & 2.11 & 2.78 & 128
    & 0.21 & 0.36 & 97 \\
    
    & \quad Crossmodal (N=4)
    & \underline{2.70} & \underline{0.99} & 128
    & \underline{4.90} & \underline{1.28} & 121
    & \underline{14.24}	& \underline{13.95}	& 182
    & \underline{2.82} & \underline{3.05} & 128
    & \textbf{0.76} & \underline{0.64} & 97 \\

    \midrule
    
    \multirow{9}{*}{\makecell{\textbf{Video}\\\textbf{Dominant}}}
    
    & Sink (N=2)
    & \textbf{5.47} & \textbf{8.54} & 144
    & \textbf{2.07} & \textbf{6.87} & 147
    & \textbf{3.57} & \textbf{3.87} & 117
    & \textbf{0.28} & \textbf{2.24} & 9
    & \underline{-0.02} & 0.00 & 29 \\
    
    & \quad Unimodal (N=2)
    & 1.93 & 3.54 & 72
    & 0.35 & 3.43 & 73
    & -0.01 & -5.00 & 58
    & 0.18 & 1.31 & 4
    & \textbf{0.00} & \textbf{0.03} & 14 \\
    
    & \quad Crossmodal (N=2)
    & \underline{3.03} & \underline{4.53} & 72
    & \underline{1.25} & \underline{4.48} & 73
    & \underline{3.53} & \underline{3.72} & 58
    & \underline{0.26} & \underline{2.19} & 4
    & \underline{-0.02} & \underline{0.01} & 14 \\
    
    \addlinespace[0.15em]
    \cmidrule(lr){2-17}
    
    & Sink (N=3)
    & \textbf{4.40} & \textbf{7.12} & 86
    & \textbf{1.62} & \textbf{5.88} & 109
    & \textbf{3.45} & \textbf{3.66} & 76
    & \underline{0.08} & \textbf{1.70} & 3
    & -0.03 & -0.06 & 16\\
    
    & \quad Unimodal (N=3)
    & 1.72 & 3.19 & 43
    & 0.31 & 3.15 & 54
    & 0.13 & -4.57 & 38
    & 0.08 & 1.44 & 1
    & \underline{0.00} & \textbf{0.06} & 8 \\
    
    & \quad Crossmodal (N=3)
    & \underline{2.15} & \underline{3.70} & 43
    & \underline{1.01} & \underline{4.11} & 54
    & \underline{3.30} & \underline{3.15} & 38
    & \textbf{0.20} & \underline{1.60} & 1
    & \textbf{0.01} & \textbf{0.06} & 8 \\
    
    \addlinespace[0.15em]
    \cmidrule(lr){2-17}
    
    & Sink (N=4)
    & \textbf{3.10} & \textbf{6.28} & 60
    & \textbf{1.10} & \textbf{4.78} & 85
    & \textbf{3.30} & \textbf{3.28} & 52
    & \underline{0.06} & \underline{1.29} & 2
    & \underline{-0.01} & \textbf{0.07} & 13 \\
    
    & \quad Unimodal (N=4)
    & 1.27 & 2.80 & 30
    & 0.24 & 2.77 & 42
    & 0.18 & -4.46 & 26
    & \textbf{0.08} & \textbf{1.33} & 1
    & -0.02 & -0.02 & 6 \\
    
    & \quad Crossmodal (N=4)
    & \underline{1.45} & \underline{3.02} & 30
    & \underline{0.63} & \underline{3.57} & 42
    & \underline{3.00} & \underline{2.56} & 26
    & 0.07 & 1.25 & 1
    & \textbf{0.02} & \underline{0.04} & 6 \\
    
    \bottomrule
    \end{tabular}
}
\vspace{-4mm}
\end{table*}

%% file: sec/5_application.tex
\section{Application: Mitigating Object Hallucination in AVLLMs}
In this section, we analyze object hallucination unique to AVLLMs and propose a mitigation method that leverages our findings on cross-modal sink tokens.

\subsection{Object Hallucination in AVLLMs}
\label{hall_attn}

\begin{figure}[t]
    \centering
    \vspace{-3mm}
    \includegraphics[width=0.9\linewidth]{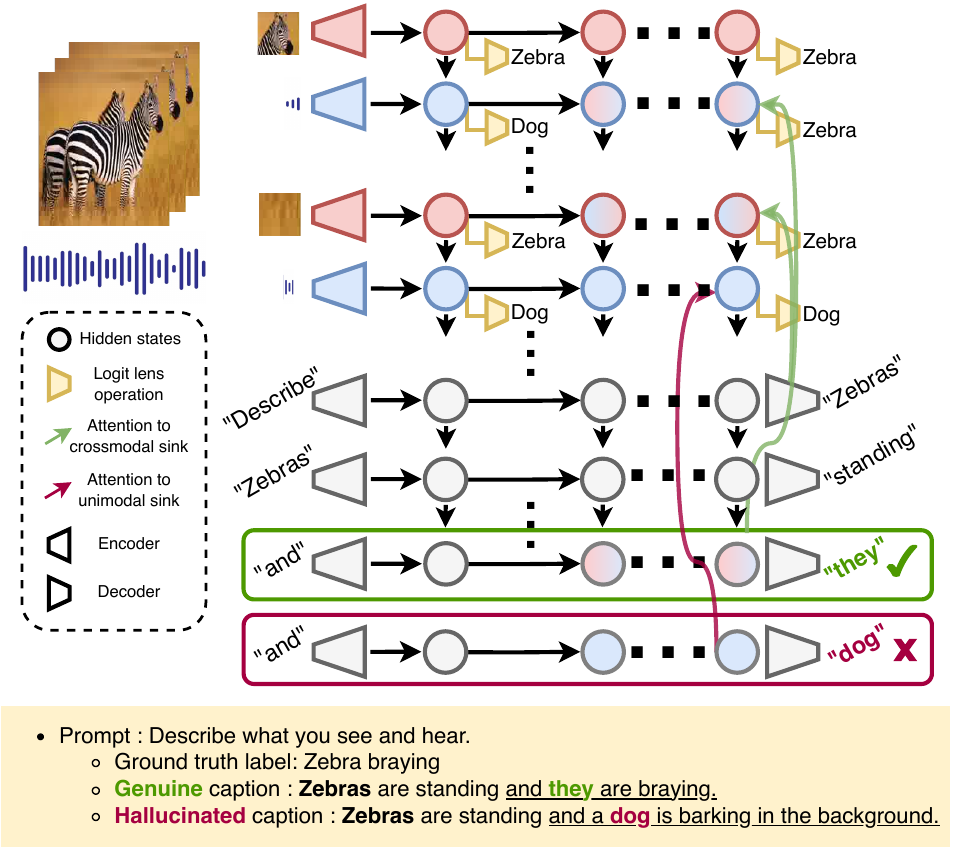}
    \vspace{-1mm}
    \caption{\textbf{Example of object hallucination in AVLLMs.} While the video modality correctly recognizes the object as a zebra, the audio modality misinterprets the zebra's braying as a dog's bark, causing the hallucinated object \emph{dog} to appear in the caption.}
    \label{fig:hall_ex}
    \vspace{-2mm}
\end{figure}
In AVLLMs, unresolved audio–visual disagreement gives rise to a new type of object hallucination.
In scenarios where one modality provides a misinterpretation of an object, this conflicting signal may persist across layers, failing to be harmonized with the correct modality. Failure to fully suppress the erroneous modality leads to a leakage of incorrect semantic cues, resulting in the model generating captions that incorporate both the correct object and the misinterpreted object. To systematically investigate this phenomenon, we leverage the VGGSound animal category, where such hallucinations frequently occur. Figure~\ref{fig:hall_ex} illustrates a representative instance: the audio modality misinterprets a zebra's braying as a dog's bark, which subsequently manifests as a hallucinated object in the final caption. Consistent with findings by \cite{nishimura2024audio}, these hallucinated objects show a tendency to emerge in the second clause, often contextualized as `background' elements.

Delving deeper into these hallucinations, we investigated the model's attention dynamics during the generation of genuine versus hallucinated objects. To enable a controlled comparison, we restricted our analysis to objects appearing in the second clause. Using the 70 samples obtained for each case, we performed a layer-wise comparison of the attention allocated to cross-modal sink and unimodal sink tokens (See Appendix~\ref{just} for the analysis on object tokens). 
\begin{figure}[t]
    \centering
    \includegraphics[width=0.9\linewidth]{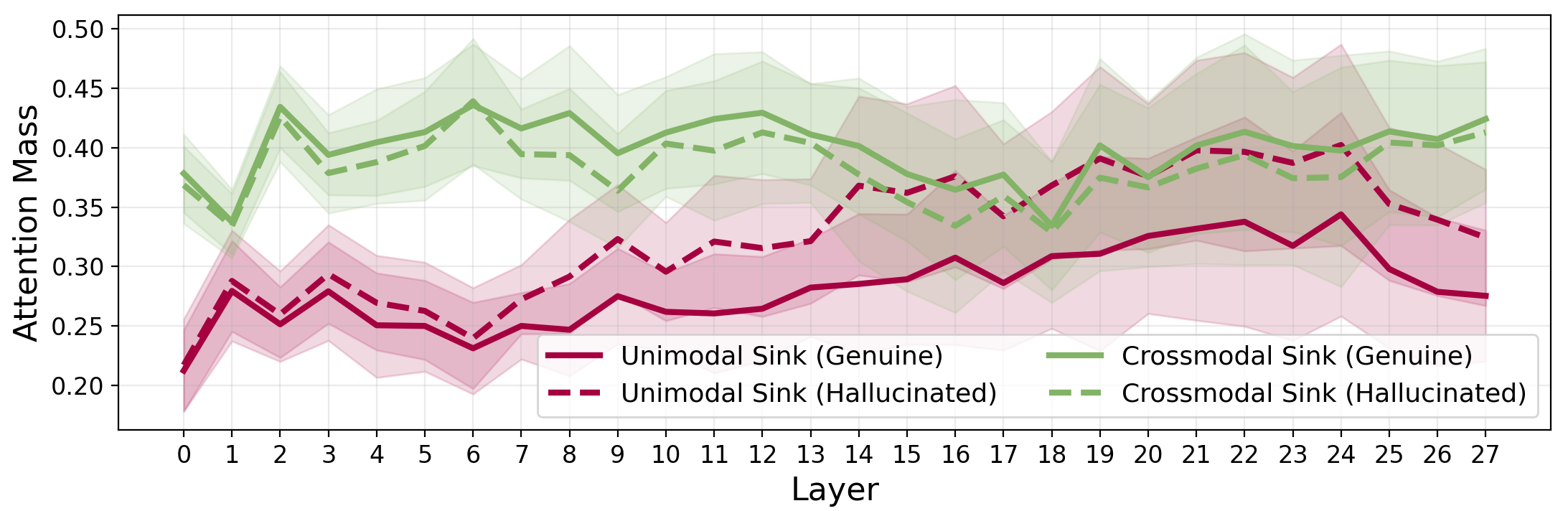}
    \vspace{-1mm}
    \caption{\textbf{Averaged attention mass to cross-modal and unimodal sink tokens across 70 genuine and 70 hallucinated samples.} Genuine object maintain dominant attention on cross-modal sinks across all layers. Conversely, hallucinated object reveal a significant surge in attention to unimodal sinks, occasionally surpassing that of cross-modal sinks.
}
    \label{fig:cross_uni}
    \vspace{-3mm}
\end{figure}
Figure~\ref{fig:cross_uni} illustrates the proportion of attention mass captured by each sink type within the total audio-visual attention. When generating genuine objects, attention to cross-modal sink tokens remains consistently higher than that to unimodal sink tokens across all layers. In contrast, for hallucinated objects, the attention directed toward unimodal sink tokens becomes more prominent, occasionally surpassing that of cross-modal
sink tokens. These results suggest that excessive attention to unimodal sinks--which are devoid of cross-modal cues--can lead to the leakage of misinterpreted information. Since these sinks do not benefit from cross-modal interaction, erroneous signals stored within them remain unresolved and subsequently manifest as a hallucinated object in the generated caption.

\subsection{Adaptive Sink-Guided Decoding} 
Building on the insight from Section~\ref{hall_attn} that improper attention allocation can lead to object hallucinations, we propose Adaptive Sink-Guided Decoding (ASD) to mitigate this issue. ASD dynamically adjusts the attention weights assigned to cross-modal and unimodal sink tokens during the generation process.

Specifically, we identify global sink tokens and categorize them into cross-modal ($\mathcal{S}_{\text{cross}}$) and unimodal ($\mathcal{S}_{\text{uni}}$) types based on their layer-averaged MDS. Subsequently, at each decoding step $t$, we perform two parallel forward passes: (i) the original model forward pass, and (ii) a \textit{calibrated} forward pass where attention to cross-modal and unimodal sink tokens is rebalanced.

In the calibrated pass, we amplify the attention toward cross-modal sink tokens while suppressing that of unimodal sink tokens as follows:
\begin{align}
\tilde{A}_{t,j} &\leftarrow {A}_{t,j} + \alpha \lvert {A}_{t,j} \rvert, \quad j \in \mathcal{S}_{\text{cross}}, \\
\tilde{A}_{t,j} &\leftarrow {A}_{t,j} - \alpha \lvert {A}_{t,j} \rvert, \quad j \in \mathcal{S}_{\text{uni}},
\end{align}
where $A_{t,j}$ denotes the raw attention weights of the token $j$ at decoding step $t$, and $\alpha$ controls the modulation magnitude, set to 0.6 in all experiments. 

We leverage the calibrated pass to steer generation, formulating decoding as an adaptive linear combination in log-probability space between the original and calibrated passes. The $t$-th token is thus sampled from the adjusted distribution:
\begin{equation}
\label{eq:guidance}
\begin{split}
\log \tilde{P}(y_t | \mathbf{x}, y_{<t}) &= \gamma_t \log P_{\text{cali}}(y_t | \mathbf{x}, y_{<t}) \\
&\quad + (1 - \gamma_t) \log P_{\text{orig}}(y_t | \mathbf{x}, y_{<t}),
\end{split}
\end{equation}
where $\mathbf{x}$ represents the multimodal input, $y_{<t}$ denotes the sequence of preceding tokens, $P_{\text{cali}}$ and $P_{\text{orig}}$ denote the probability distributions obtained from the calibrated and original forward passes, respectively, and $\gamma_t$ serves as the adaptive guidance scale. 

Crucially, $\gamma_t$ is determined adaptively at each step to prioritize the calibrated distribution when the risk of hallucination is high. We identify such ``hallucination-prone" steps by detecting excessive attention to unimodal sinks relative to cross-modal sinks in the original forward pass. Accordingly, we compute the base guidance scale $\gamma_{t}^{\text{base}}$ as the proportion of total sink attention attributed to unimodal sinks:
\begin{equation}
\gamma_{t}^{\text{base}} = \frac{\bar{A}_{t, \text{uni}}}{\bar{A}_{t,\text{uni}} + \bar{A}_{t,\text{cross}}},
\end{equation}
where $\bar{A}_{t, \text{uni}}$ and $\bar{A}_{t, \text{cross}}$ denote the average attention received by unimodal and cross-modal sink tokens, respectively, at decoding step $t$. Finally, to ensure stability and prevent abrupt shifts, we apply quadratic soft gating and momentum-based temporal smoothing to $\gamma_{t}^{\text{base}}$ to obtain final guidance scale $\gamma_t$ (see Appendix~\ref{appendix:smoothing} for details).

\subsection{Experimental Setting}
To assess the efficacy of our approach, we conduct experiments using two representative AVLLMs from Qwen and video-SALMONN families: Qwen2.5-Omni (7B) and video-SALMONN-o1 (7B).

\textbf{Datasets.}
We evaluate our methods on three datasets. First, to verify whether our approach successfully mitigates the targeted hallucination cases, we use the clean VGGSound-Animal subset, comprising 360 samples, introduced in \cref{hall_attn}. Additionally, to verify that our method maintains performance on general benchmarks, we evaluate on the standard VGGSound and AudioSet~\cite{gemmeke2017audio} datasets. AudioSet is a large-scale audio-visual dataset containing video clips with labeled sound events. For these benchmarks, we utilize approximately 1,000 and 700 clean samples, respectively (see Appendix ~\ref{subsubsec:dataset_hall} for details).
As the annotations in these datasets are limited to the primary sound source, we augment the ground truth using an object detection model~\cite{carion2020end} to identify additional visible entities. We thus establish the ground truth by combining the labeled source with these detected visible objects.

\textbf{Evaluation metrics.}
To evaluate object hallucination in the captioning task, we adopt the CHAIR~\cite{rohrbach2018object} metrics, which measure the proportion of mentioned objects that are absent in the ground-truth annotations. CHAIR reports sentence-level ($\mathrm{C}_S$) and instance-level ($\mathrm{C}_I$) hallucination rates, defined as
\begin{equation}
    C_I = \frac{|\textrm{\scriptsize \{hallucinated objects\}}|}{|\textrm{\scriptsize \{all mentioned objects\}}|}, C_S = \frac{|\textrm{\scriptsize \{captions w/ hallucinated objects\}}|}{|\textrm{\scriptsize \{all captions\}}|}, \nonumber
\end{equation}

Given that the standard CHAIR is restricted to fixed MS COCO~\cite{lin2014microsoft} objects, we extend its object vocabulary and synonym lists to align with the specific taxonomy of each dataset.  Following~\cite{liu2024payingattentionimagetrainingfree}, we report the F1 score to evaluate both the richness and accuracy of generated descriptions.

We further adopt ALOHa~\cite{petryk2024aloha}, an open-vocabulary hallucination evaluation metric to overcome the closed-vocabulary limitation of CHAIR.
ALOHa utilizes a large language model (GPT-3.5-turbo) to extract groundable objects from generated captions and computes their semantic similarity to the ground-truth objects derived from reference captions.

\paragraph{Baselines.}
We compare our approach with two training-free hallucination mitigation methods adapted to the audio–visual setting (see Appendix \ref{appendix:baseline_details} for adaptation details): 
(i) PAI~\cite{liu2024payingattentionimagetrainingfree}, which modulates the attention mechanism to amplify the contribution of multimodal tokens; (ii) VCD~\cite{leng2023mitigatingobjecthallucinationslarge}, which contrasts the output logits derived from original inputs with those from distorted counterparts.
\input{table/5.3_result}
\begin{figure}[t]
\vspace{-3mm}
    \centering
    \begin{minipage}{0.48\columnwidth}
        \centering
         \includegraphics[width=\linewidth]{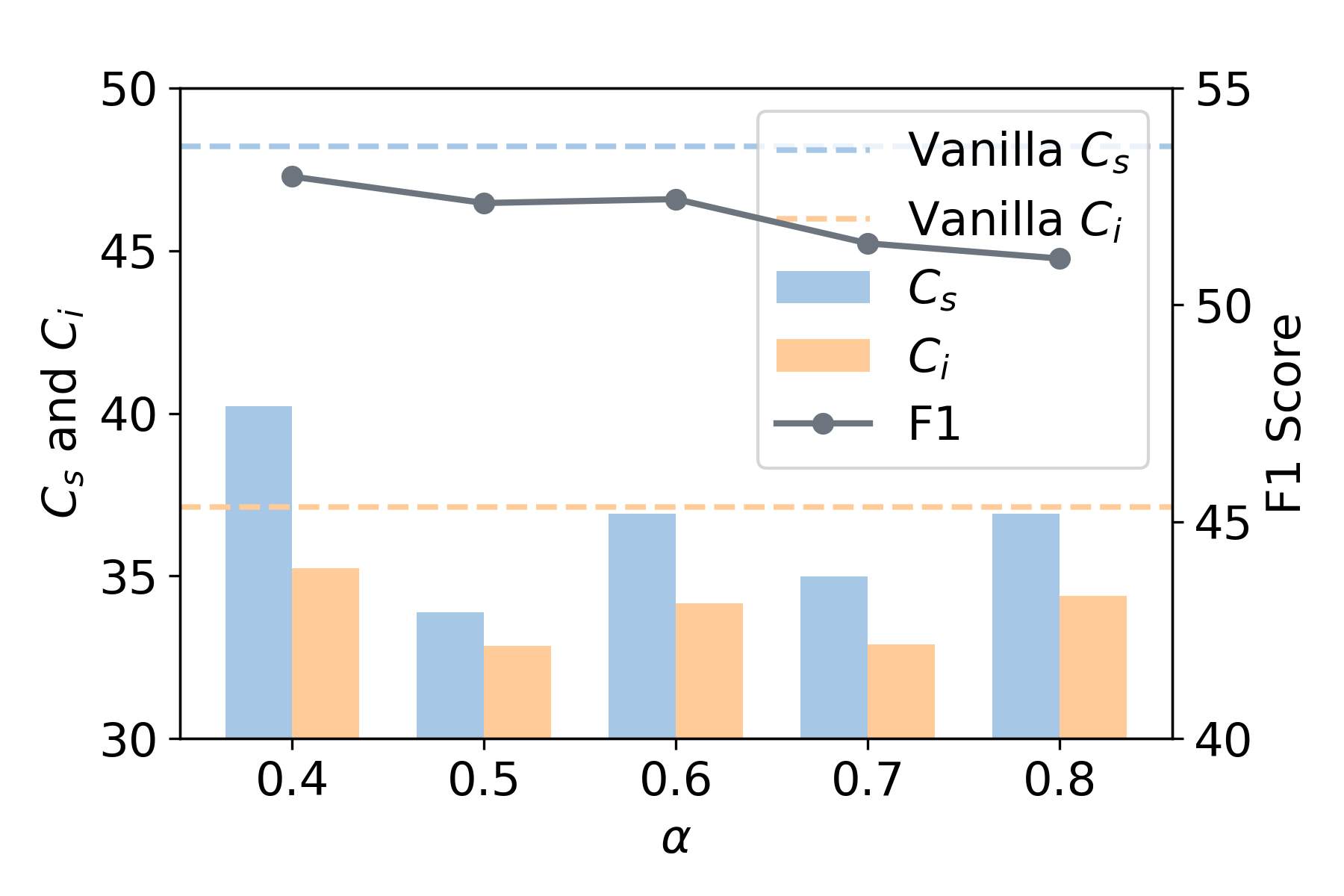}
         \vspace{-7mm}
        \caption*{(a) Qwen2.5-Omni(7B)}
    \end{minipage}
    \hfill
    \begin{minipage}{0.48\columnwidth}
        \centering
        \includegraphics[width=\linewidth]{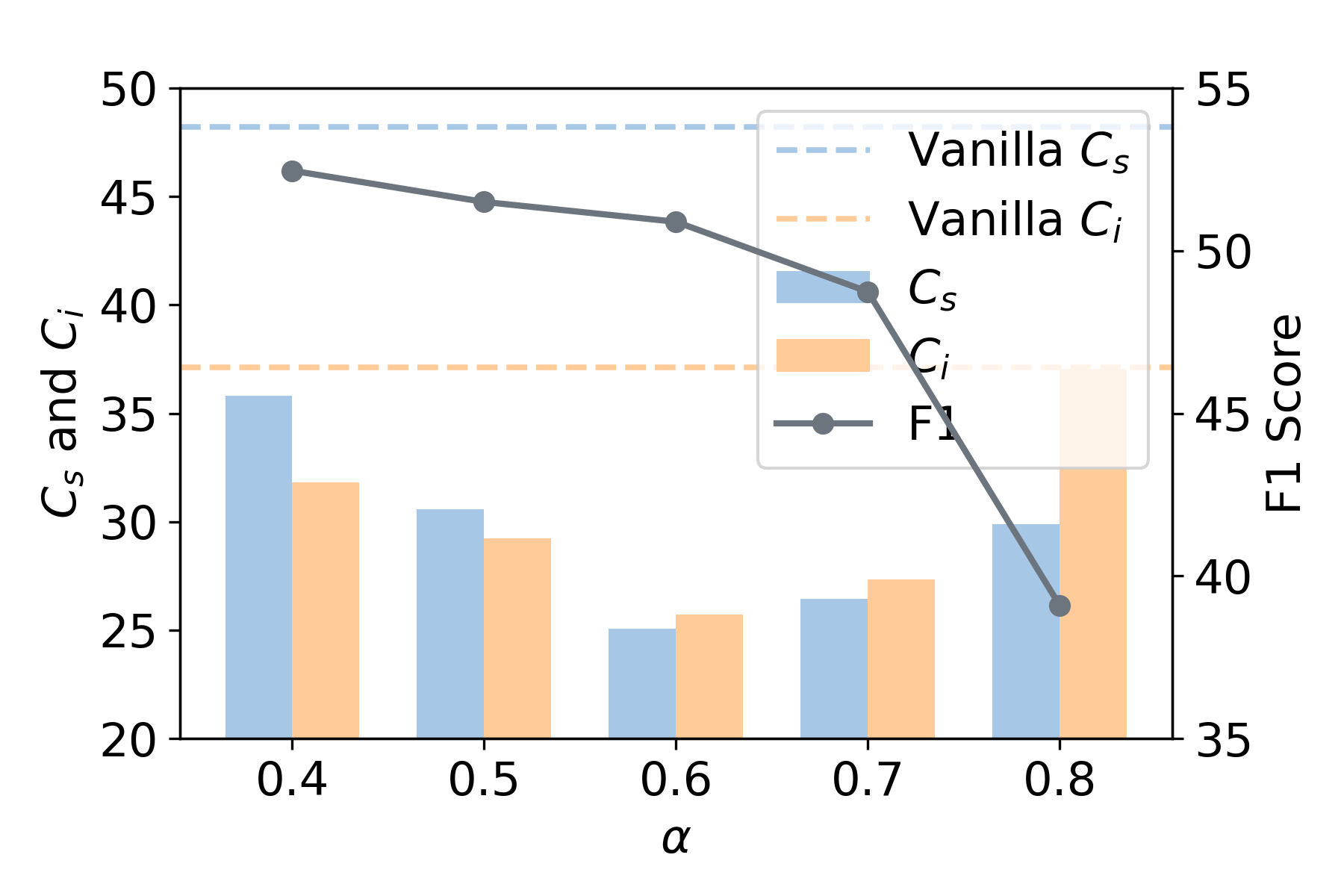}
        \vspace{-7mm}
        \caption*{(b) video-SALMONN-o1(7B)}
    \end{minipage}
    \vspace{-1mm}
    \caption{
Parameter sensitivity of $\alpha$ with CHAIR metrics.
}
    \label{fig:hall_ablation}
    \vspace{-6mm}
\end{figure}

\subsection{Experimental Results} 
\cref{tab:hallucination} presents the results. Whereas PAI and VCD offer negligible gains or occasionally exacerbate hallucinations, ASD delivers consistent and substantial reductions in hallucinations. Notably, our improvements are maximized on the VGGSound-Animal benchmark-where AVLLM-specific hallucinations induced by audio-visual disagreement are most prevalent-validating our targeted approach. Crucially, ASD maintains this superiority on the more general VGGSound-All and AudioSet datasets, underscoring its robustness.

\textbf{Ablation study on $\alpha$.} \cref{fig:hall_ablation} illustrates the impact of the modulation magnitude $\alpha$ on the VGGSound-Animal dataset. We observe that ASD is robust to hyperparameter variations, consistently reducing hallucinations across a wide range of values. That said, we note a trade-off: insufficient $\alpha$ yields limited hallucination mitigation, while excessive $\alpha$ degrades the richness of the generated captions. Thus, selecting an optimal $\alpha$ is essential to balance hallucination suppression with the maintenance of details. See Appendix~\ref{app:obj_hall_red} for additional analyses and Appendix~\ref{app:qualitative_results} for qualitative results.

%% file: table/5.3_result.tex
\begin{table}[t]
\centering
\setlength{\tabcolsep}{3pt}
\caption{ \textbf{Quantitative results of ASD.} We evaluate hallucination using ALOHa and CHAIR, and assess caption richness using F1 scores.}
\vspace{-1mm}

\label{tab:hallucination}
\resizebox{\linewidth}{!}{%
   \begin{tabular}{ l l 
    c c c >{\columncolor{gray!15}}c 
    c c c >{\columncolor{gray!15}}c }
    
    \toprule
    \textbf{Dataset} & \textbf{Method}
    & \multicolumn{4}{c}{\textbf{Qwen2.5-Omni(7B)}}
    & \multicolumn{4}{c}{\textbf{video-SALMONN-o1 (7B)}} \\
    \cmidrule(lr){3-6}\cmidrule(lr){7-10}
    &
    & ALOHa $\uparrow$ & Cs $\downarrow$ & Ci $\downarrow$ & F1 $\uparrow$
    & ALOHa $\uparrow$ & Cs $\downarrow$ & Ci $\downarrow$ & F1 $\uparrow$ \\
    \midrule

    \multirow{4}{*}{\shortstack[l]{\textbf{VGGSound-}\\\textbf{Animal}}}
    & Vanilla
    & 40.71 & 48.21 & 37.13 & 55.24
    & 36.21 & 37.74 & 32.09 & 53.68 \\
    & PAI
    & 39.52 & 51.24	&	38.11	&	55.11
    & 36.99& 35.26	&	31.18	&	53.16 \\
    & VCD
    & 40.27 & 51.52 & 41.28 & 52.43
    & 36.40 & 39.39	&	33.40	&	53.37\\
    & \textbf{ASD}
    & \textbf{42.77} & \textbf{36.91} & \textbf{34.15} & 52.44
    & \textbf{43.29 }& \textbf{25.07} & \textbf{25.71} & 50.89 \\
    \midrule

    \multirow{4}{*}{\shortstack[l]{\textbf{VGGSound-}\\\textbf{All}}}
    & Vanilla
    & 35.02 & 30.70 & \textbf{20.67} & 58.69
    & 32.74 & 30.63 & 22.39 & 53.40 \\
   
    & PAI
    & 34.68 & 32.21	&	21.52	&	58.47
    & 32.44 & 29.29	&	22.01	&	53.15 \\
     & VCD
    & 34.60 & 32.63 & 22.36 & 57.09
    & 30.28 & 30.76	&	24.31&		50.02 \\
    & \textbf{ASD}
    & \textbf{38.89} & \textbf{29.65} & 21.74 & 55.81
    & \textbf{36.63} & \textbf{21.11} & \textbf{18.42} & 50.10 \\
    \midrule

    \multirow{4}{*}{\textbf{Audioset}}
    & Vanilla
    & 38.24 & 8.92 & 10.93 & 69.73
    & 36.81 & 11.39 & 14.91 & 67.27 \\
    
    & PAI
    & 36.94 &11.84	&	13.09	&	73.22
    & 36.05 & 10.95	&	14.54	&	67.64\\
    & VCD
    & 36.98 & 12.28 & 14.88 & 71.12
    & 32.50 & 9.34	&	12.52	&	67.74 \\
    & \textbf{ASD}
    & \textbf{38.32} & \textbf{8.54} & \textbf{10.20} & 72.98
    & \textbf{39.64} & \textbf{6.57} & \textbf{9.50} & 67.29 \\
    \bottomrule
    \end{tabular}%
}
\vspace{-2mm}
\end{table}

%% file: sec/6_conclusion.tex
\section{Conclusion}
In this paper, we investigated the internal mechanisms of AVLLMs to understand where cross-modal information is stored in each modality tokens.
To enable tracing of bidirectional information flow between audio and video modalities, we adapt causal tracing via a unimodal dominance framework.
Using this framework, we showed that cross-modal information is primarily localized in cross-modal sink tokens.
Building on these insights, we proposed a simple, training-free object hallucination mitigation method that steers generation toward cross-modal sink tokens.
We hope our findings provide foundational insights for future work on interpreting, diagnosing, and improving AVLLMs.

%% file: sec/appendix.tex
\newpage
\appendix
\onecolumn

\section*{Appendix Outline}
\vspace{0.5em}
\hrule
\vspace{1em}

\noindent The appendix provides detailed implementations, additional analyses, and qualitative examples supporting the main paper. The structure is organized as follows:

\vspace{1em}

\noindent \textbf{\large \ref{sec:imp_details}. Implementation Details}
\begin{itemize}
    \item \textbf{\ref{subsec:sink_tokens}. Sink Tokens}
    \begin{itemize}
        \item \ref{app:definition_of_sink_tokens}. Definition of Sink Tokens
        \item \ref{subsubsec:sink_dim}. Selection of Sink Dimension
    \end{itemize}
    
    \item \textbf{\ref{subsec:dataset}. Dataset}
    \begin{itemize}
        \item \ref{appendix:dataset_for_causal}. Datasets for Causal Tracing
        \item \ref{subsubsec:dataset_hall}. Datasets for Object Hallucination Mitigation Evaluation
    \end{itemize}
    
    \item \textbf{\ref{subsec:algo}. Adaptive Sink-Guided Decoding Algorithm}
    \begin{itemize}
        \item \ref{just}. Insignificance of Object Tokens in Hallucination Detection
        \item \ref{appendix:smoothing}. Details of Stability Functions
    \end{itemize}
    
    \item \textbf{\ref{appendix:baseline_details}. Details of Adapting Baseline Approaches to Audio-Visual Setting}
\end{itemize}

\vspace{0.5em}

\noindent \textbf{\large \ref{sec:add_analysis}. Additional Analysis}
\begin{itemize}
    \item \textbf{\ref{appendix:causal_tracing_further}. Causal Tracing Analysis}
    \begin{itemize}
        \item \ref{appendix:bottom_up}. Bottom-up Analysis
        \item \ref{appendix:patching_location}. Patching Locations Analysis
        \item \ref{app:causl_tracing_ablation}. Causal Tracing Results with Different Patching Locations
        \item \ref{subsubsec:layerwise}. Layerwise Analysis
        \item \ref{app:corruption_alt}. Alternative Corruption Methods Analysis
        \item \ref{app:mds_stat}. MDS Statistics
    \end{itemize}
    
    \item \textbf{\ref{app:obj_hall_red}. Object Hallucination Reduction}
    \begin{itemize}
        \item \ref{addition:object_hall_case}. Object Hallucination Case Analyses
        \item \ref{app:add_models}. Additional Results on Different Models
        \item \ref{app:reverse_asd}. Reverse ASD as a Counterfactual Test
        \item \ref{app:av_decoding_baselines}. Comparison with Inference-Time Interventions in AVLLMs
        \item \ref{app:overhead}. Latency Overhead of ASD
    \end{itemize}
\end{itemize}

\vspace{0.5em}

\noindent \textbf{\large \ref{app:qualitative_results}. Qualitative Results}

\vspace{0.5em}

\noindent \textbf{\large \ref{app:limit}. Limitations}

\vspace{1em}
\hrule
\newpage

\section{Implementation Details}
\label{sec:imp_details}
\subsection{Sink Tokens}
\label{subsec:sink_tokens}
\subsubsection{Definition of sink tokens}
\label{app:definition_of_sink_tokens}
Sink tokens are tokens that receive disproportionately high attention, 
exhibit abnormally large activations in specific hidden dimensions 
(so-called \emph{sink dimensions} $\mathcal{D}_{\text{sink}}$)~\cite{kang2025toldvisualattentionsink}, 
and have high feature norms. 
Prior studies have characterized sink tokens using diverse criteria, 
including attention aggregation patterns and activation-based signatures~\citep{darcetvision,kang2025toldvisualattentionsink,sun2024massive}.

Following~\cite{kang2025toldvisualattentionsink}, we identify sink tokens at each layer 
based on their activations in predefined sink dimensions. 
Formally, for each layer $l$, the set of sink tokens is defined as
\begin{equation}
\label{eq_sink}
\hat{\mathcal{I}}^{l} = \left\{ j \in \mathcal{I} \,\middle|\, \phi\left(\mathbf{x}^{l-1}_{j}\right) \geq \tau \right\},
\end{equation}
where $\mathbf{x}^{l-1}_{j}$ denotes the pre-attention hidden representation of token $j$ at layer $l$, 
$\tau$ is a predefined threshold, and $\phi(\cdot)$ is a sink characteristic function defined as
\begin{equation}
\phi(\mathbf{x}^{l-1}_{j}) 
= \max_{d \in \mathcal{D}_{\text{sink}}} 
\left| \mathrm{RMSNorm}(\mathbf{x}^{l-1}_{j})[d] \right|.
\end{equation}
Here, $\mathrm{RMSNorm}(\cdot)$ denotes root-mean-square normalization, 
and $\mathcal{D}_{\text{sink}}$ is the set of sink dimensions determined from the backbone LLM.

\begin{wrapfigure}{r}{0.35\linewidth}
    \centering
    \vspace{-10pt}  
    \includegraphics[width=0.95\linewidth]{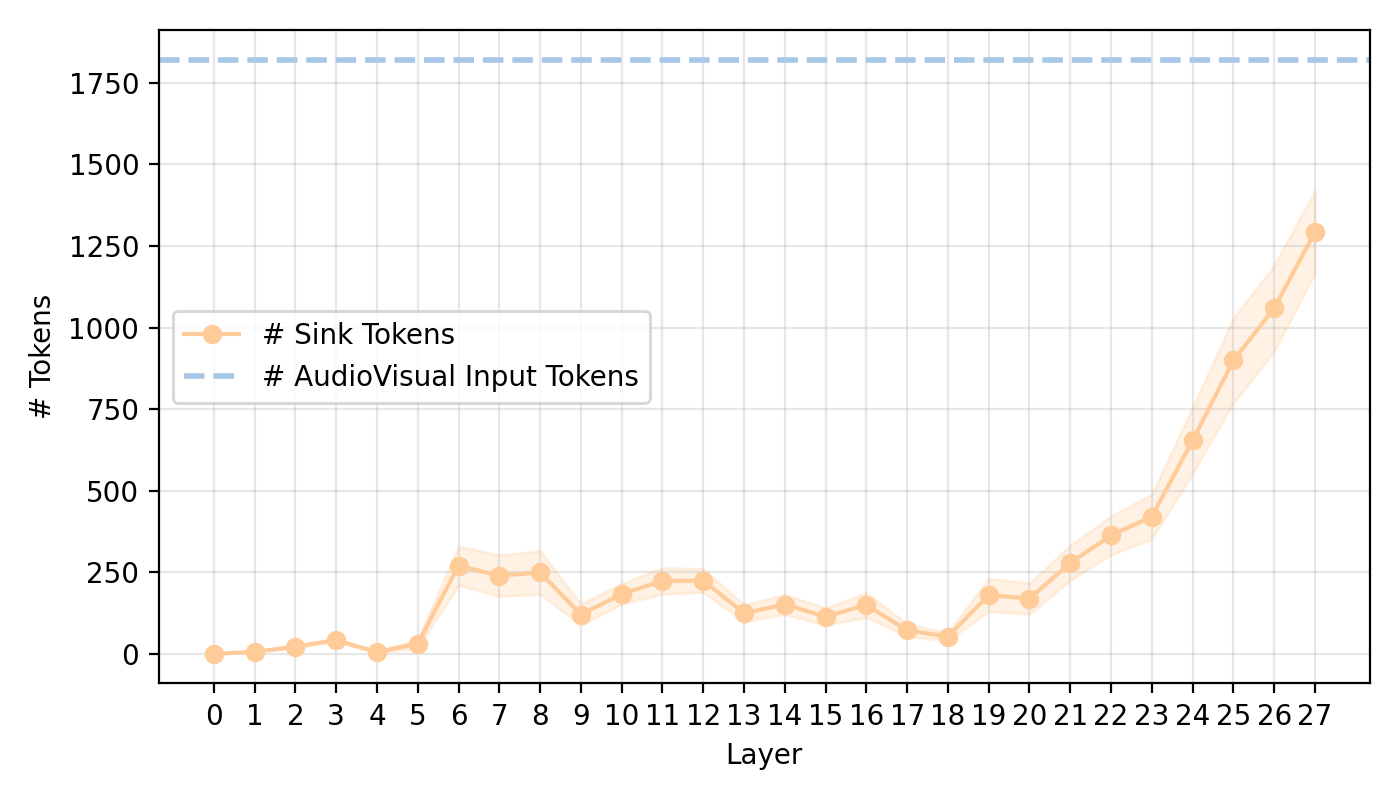} 
    \caption{Layer-wise average number of sink tokens across 100 samples in Qwen2.5-Omni (7B). The shaded region denotes one standard deviation. The number of sink tokens vary substantially across layers.}
    \label{fig:layerwise_sink_num}
    \vspace{-10pt}
\end{wrapfigure}

\cite{kang2025toldvisualattentionsink} report that visual sink tokens in VLMs emerge in early layers 
and persist across deeper layers. 
In contrast, we observe substantial variability in sink token identities across layers in audio-visual LLMs. 
Figure~\ref{fig:layerwise_sink_num} shows the layer-wise average number of sink tokens in Qwen2.5-Omni (7B). 
The number of sink tokens varies significantly across layers and increases sharply in deeper layers, 
where a large fraction of tokens become classified as sink tokens. 
This suggests that sink behavior in AVLLMs is highly dynamic and layer-dependent, 
likely reflecting hierarchical multimodal information aggregation.

Motivated by this observation, we introduce a \emph{global sink token} definition that accounts for cross-layer variability. 
Instead of using layer-wise sink sets $\hat{\mathcal{I}}^{l}$, 
we measure how frequently each token is classified as a sink token across layers. 
Specifically, for each token $j$, we define its sink frequency score as
\begin{equation}
s_j = \sum_{l=1}^{L} \mathbb{I}\left[j \in \hat{\mathcal{I}}^{l}\right],
\end{equation}
where $L$ is the total number of layers and $\mathbb{I}(\cdot)$ is the indicator function.

We then sort tokens by $s_j$ and define the top-$K$ tokens as \emph{global sink tokens}, 
where $K$ is set to a fixed fraction of the maximum input length:
\begin{equation}
\mathcal{I}_{\text{global}} = \operatorname{TopK}\left(\{s_j\}_{j \in \mathcal{I}},\, K = \frac{|\mathcal{T}|}{N}\right),
\end{equation} 
with $|\mathcal{T}|$ denoting the input token length and $N$ a normalization constant.
This definition yields a stable set of sink tokens that persist across layers and avoids over-selecting transient layer-specific sinks.

\subsubsection{Selection of sink dimension}
\label{subsubsec:sink_dim}
\begin{table}[t]
\centering
\scriptsize
\caption{Model-specific sink dimensions $\mathcal{D}_{\text{sink}}$ and thresholds $\tau$ used to identify layer-wise sink tokens.}
\label{tab:sink_config}
\setlength{\tabcolsep}{6pt}
\begin{tabular}{lcc}
\toprule
\textbf{Model} & $\boldsymbol{\mathcal{D}_{\text{sink}}}$ & $\boldsymbol{\tau}$ \\
\midrule
Qwen2.5-Omni (7B)      & $(458,\;2570)$ & 25 \\
Qwen2.5-Omni (3B)      & $(318,\;1874)$ & 20 \\
video-SALMONN-o1 (7B)  & $(458,\;2570)$ & 25 \\
video-SALMONN2+ (7B)   & $(458,\;2570)$ & 25 \\
video-SALMONN2+ (3B)   & $(318,\;1874)$ & 20 \\
\bottomrule
\end{tabular}
\end{table}

\cite{kang2025toldvisualattentionsink} reported that, even after multimodal fine-tuning, 
the dimensions exhibiting massive activations remain largely consistent with those of the base LLM. 
Specifically, visual sink tokens in VLMs exhibit massive activation along the same dimensions as the BOS token in the base LLM.

Following this observation, we select sink dimensions based on BOS-token activations in the backbone LLM.
For Qwen2.5-Omni (7B/3B) and video-SALMONN2+ (7B/3B), whose backbones are Qwen2.5-VL (7B/3B), 
we identify the BOS token’s massive-activation dimensions in the corresponding base model 
and use these dimensions as $\mathcal{D}_{\text{sink}}$ for all subsequent analyses.
For video-SALMONN-o1, which is based on Qwen2-7B, we directly adopt the sink-dimension set 
$\mathcal{D}_{\text{sink}}=\{458,\,2570\}$ reported in~\cite{kang2025toldvisualattentionsink}.
Table~\ref{tab:sink_config} summarizes the sink dimensions and thresholds used in this work.

\begin{figure}[t]
    \centering
    \hspace*{\fill}
    \begin{subfigure}{0.32\linewidth}
        \centering
        \includegraphics[width=\linewidth]{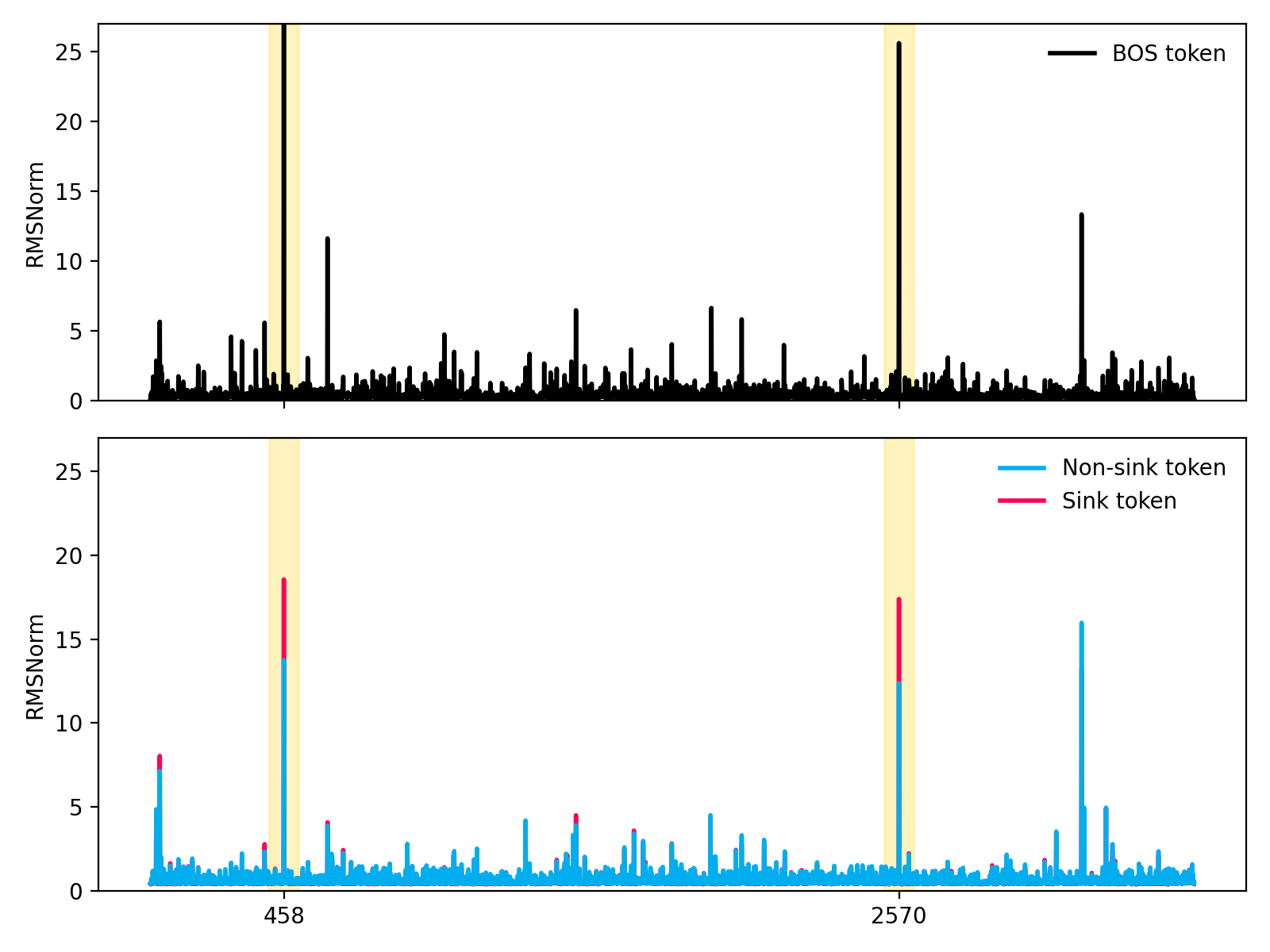}
        \caption{Qwen2.5-Omni (7B)}
        \label{fig:qwen7b_sink_dim}
    \end{subfigure}
    \hspace{0.04\linewidth}
    \begin{subfigure}{0.32\linewidth}
        \centering
        \includegraphics[width=\linewidth]{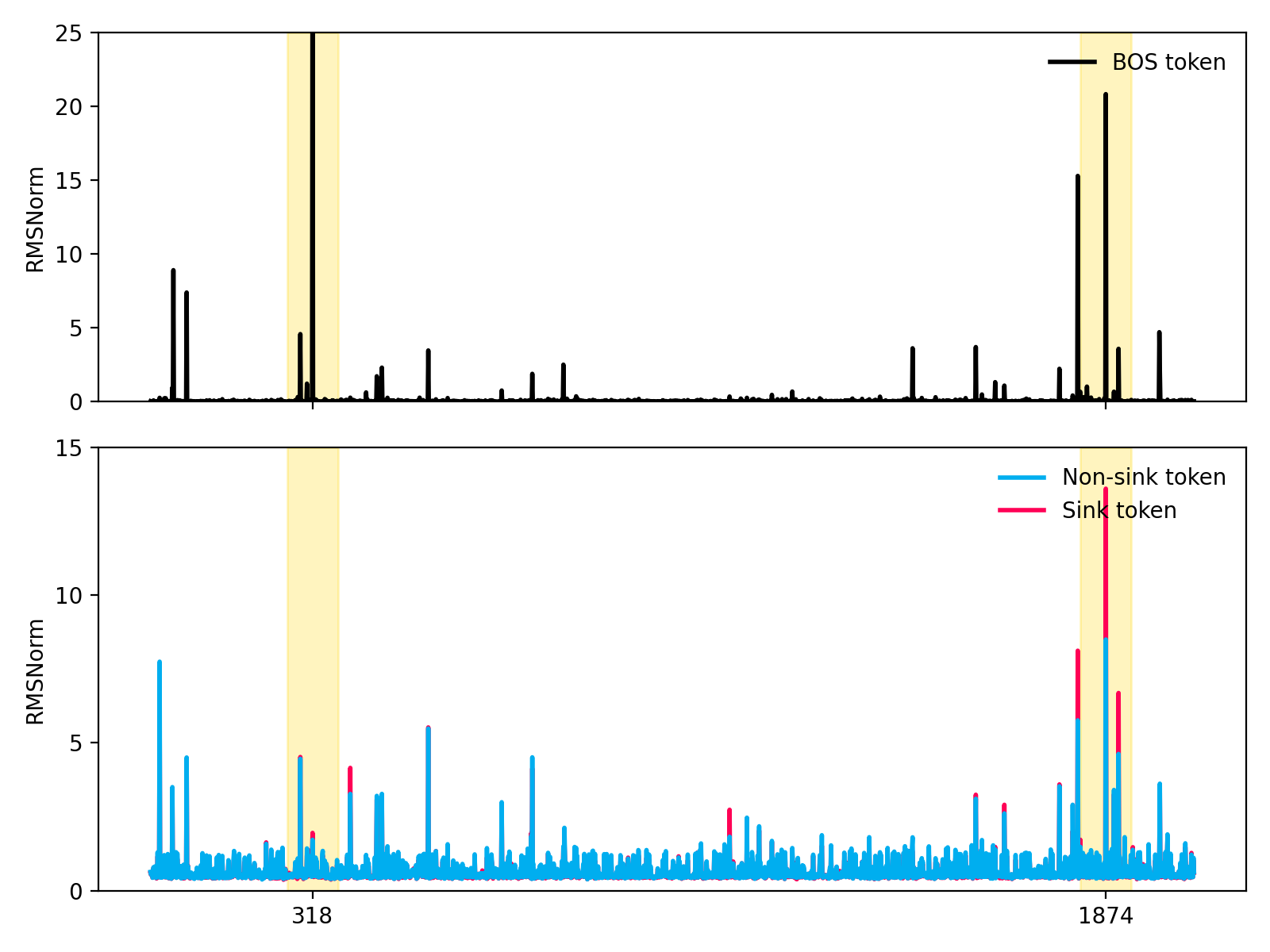}
        \caption{Qwen2.5-Omni (3B)}
        \label{fig:qwen3b_sink_dim}
    \end{subfigure}
    \hspace*{\fill}

    \vspace{0.5em}

    \begin{subfigure}{0.32\linewidth}
        \centering
        \includegraphics[width=\linewidth]{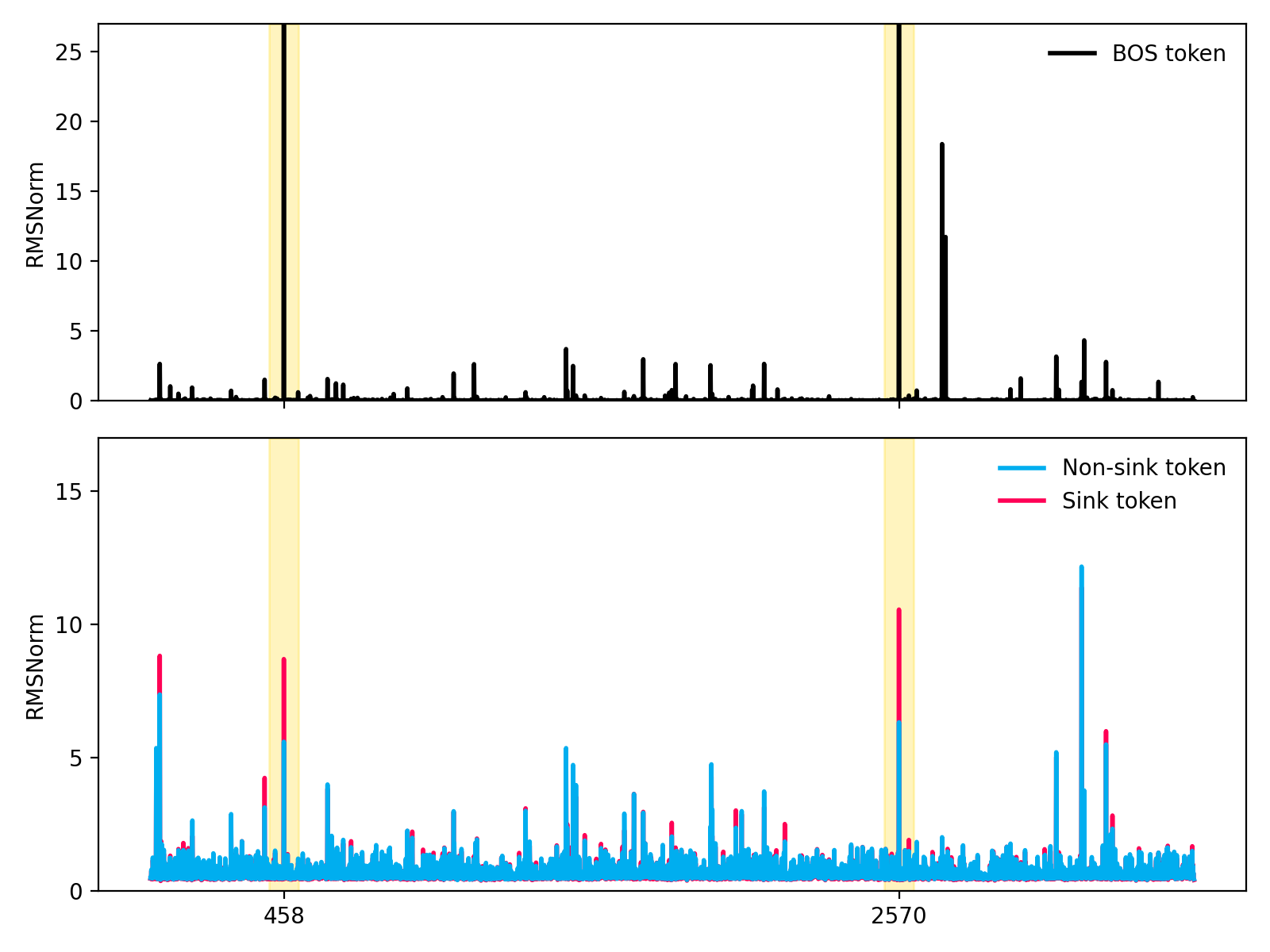}
        \caption{video-SALMONN-o1 (7B)}
        \label{fig:o17b_sink_dim}
    \end{subfigure}
    \hfill
    \begin{subfigure}{0.32\linewidth}
        \centering
        \includegraphics[width=\linewidth]{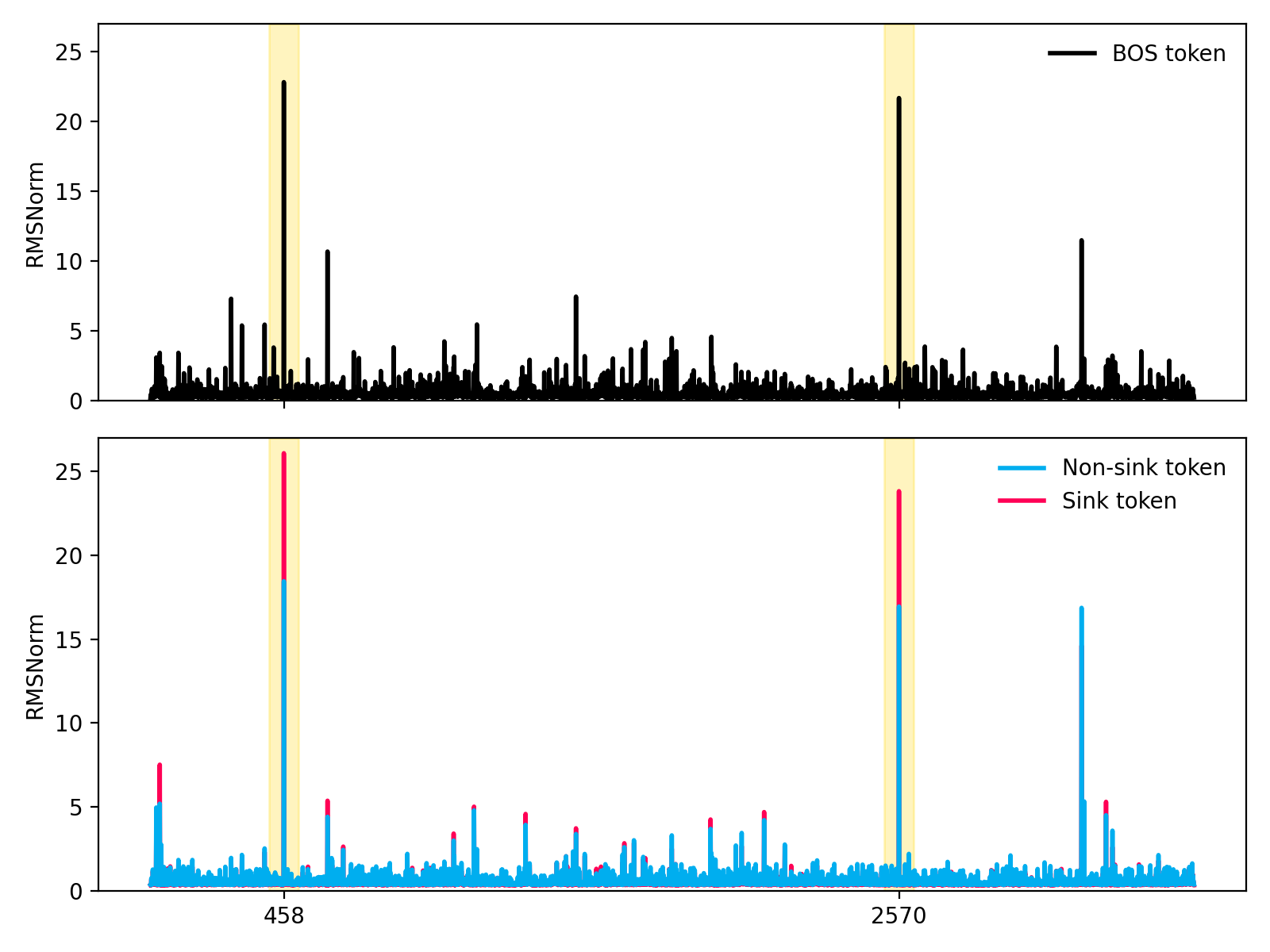}
        \caption{video-SALMONN2+ (7B)}
        \label{fig:plus7b_sink_dim}
    \end{subfigure}
    \hfill
    \begin{subfigure}{0.32\linewidth}
        \centering
        \includegraphics[width=\linewidth]{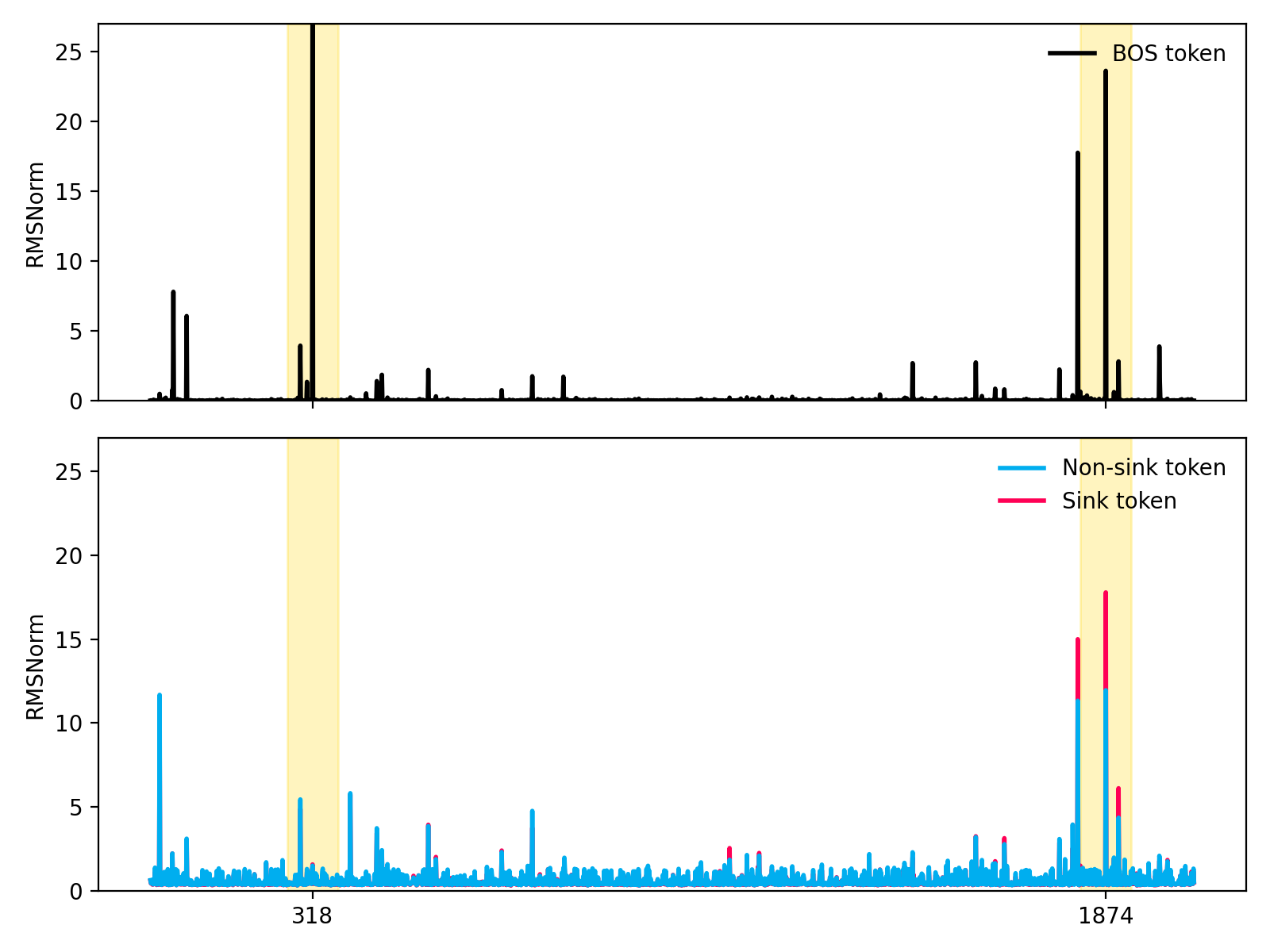}
        \caption{video-SALMONN2+ (3B)}
        \label{fig:plus3b_sink_dim}
    \end{subfigure}

    \caption{
    Dimension-wise RMSNorm magnitudes for BOS, sink, and non-sink tokens.
    Top: Qwen-based models. Bottom: SALMONN-based models.
    }
    \label{fig:sink_dim_compare}
\end{figure}

Figure~\ref{fig:sink_dim_compare} compares the dimension-wise RMSNorm magnitudes for BOS, sink, and non-sink tokens, 
averaged over 100 samples. Sink tokens consistently exhibit larger activations along the identified sink dimensions, 
validating our sink-dimension selection strategy across models.

\subsection{Dataset}
\label{subsec:dataset}
\subsubsection{Datasets for causal tracing}
\label{appendix:dataset_for_causal}

To analyze the localization of cross-modal information via causal tracing, we
construct evaluation subsets based on the \emph{Audio Dominant} and
\emph{Video Dominant} cases defined in \cref{sec:av_conflict}.
Following the observations of \cite{jiang2025bridgingearseyesanalyzing}, audio-dominant
cases predominantly correspond to human actions with distinctive acoustic cues,
whereas video-dominant cases cover sports and visually salient activities.

Based on these findings, we curate two disjoint sets of categories from the
VGGSound dataset.
For each case, we first collect representative categories explicitly mentioned
in prior work and then expand them with semantically aligned categories from
VGGSound.
To avoid ambiguity in dominance attribution, we exclude overlapping or
semantically redundant categories.
For example, categories such as \emph{people eating} and \emph{eating with
cutlery} are treated as duplicates, and only one is retained.

\paragraph{Audio dominant categories.}
The final set of audio-dominant categories consists of the following 20 classes:
people farting, people finger snapping, people slapping, people battle cry,
civil defense siren, people sobbing, people whispering, people humming,
people shuffling, police radio chatter, people burping, people nose blowing,
smoke detector beeping, people booing, people whistling, people eating,
people gargling, people coughing, people hiccup, and francolin calling.

\paragraph{Video dominant categories.}
The final set of video-dominant categories consists of the following 20 classes:
playing volleyball, playing darts, playing table tennis, skiing, striking bowling,
skateboarding, rope skipping, golf driving, playing tennis, mouse clicking,
basketball bounce, swimming, slot machine, sailing, playing squash, playing hockey,
shooting football, playing badminton, bouncing on trampoline, and parrot talking.

Each category in the VGGSound test set contains 50 samples, resulting in
1,000 samples for the audio-dominant set and 1,000 samples for the video-dominant
set before filtering.
As discussed earlier, we further filter these samples based on model-specific
prediction patterns.
Specifically, we retain only samples satisfying
$\hat{y}_{av} = \hat{y}_a \neq \hat{y}_v$ for the audio-dominant case and
$\hat{y}_{av} = \hat{y}_v \neq \hat{y}_a$ for the video-dominant case.
These filtered subsets are used for causal tracing analysis for each model.

\input{table/appendix_dataset_stats}
\cref{tab:appendix_dataset_stats} summarizes the number of samples retained
for each model after applying the dominance-based filtering.

\subsubsection{Datasets for Object Hallucination Mitigation Evaluation}
\label{subsubsec:dataset_hall}
Unlike image captioning, where dedicated benchmarks for measuring object hallucination exist (e.g., MS-COCO~\cite{lin2014microsoft}), 
there is no established benchmark for object hallucination evaluation in audio-visual captioning. 
To construct a comparable evaluation framework, we leverage audio-visual classification datasets, VGGSound and AudioSet, 
where audio labels are reliably annotated but visual background objects are not exhaustively labeled. 
To obtain comprehensive ground-truth object sets, we additionally apply object detection models to identify unlabeled visual objects.

We evaluate our method on three datasets. 
Our empirical analysis reveals that object hallucinations occur most frequently in animal-related scenarios, 
where audio-visual correlations are strong but visually ambiguous. 
Therefore, we first evaluate on a curated VGGSound-Animal benchmark to quantify hallucination mitigation in challenging cases. 
To assess generalization and ensure that our method does not degrade performance on broader distributions, 
we additionally evaluate on VGGSound-All and AudioSet.

\paragraph{VGGSound-Animal.}
We construct a curated VGGSound-Animal benchmark based on VGGSounder~\cite{zverev2025vggsounderaudiovisualevaluationsfoundation}, 
a comprehensively re-annotated multi-label evaluation set derived from the original VGGSound dataset. 
We focus on animal-related classes and exclude bird categories due to severe label ambiguity and semantic overlap. 
We further filter samples by removing instances where the VGGSounder annotations and the original VGGSound labels are inconsistent. 
From the remaining samples, we select clips in which the target animal class is present in both audio and visual modalities and is the only annotated class in the clip. 
This results in approximately 360 clean single-label animal samples.

\paragraph{VGGSound-All.}
To evaluate generalization beyond single-label animal scenarios, we construct a VGGSound-All benchmark that includes multi-label samples. 
We retain only samples where the VGGSounder annotations are consistent with the original VGGSound labels. 
For each object class, we randomly sample up to 10 clips to balance the class distribution, resulting in approximately 1,200 samples. 
Multi-label annotations from VGGSounder are treated as ground-truth object sets.

\paragraph{AudioSet.}
Given that AudioSet contains substantial annotation noise and weak cross-modal alignment, we adopt the subset curated by \cite{chen2025modalitiesequalothersdecoding}. This data was obtained through a rigorous filtering process involving ontology pruning, automated cross-modal verification, and human inspection. In total, we employ approximately 680 clean samples.

\paragraph{Ground-truth object sets.}
Following~\cite{petryk2024aloha}, all video clips are truncated to a maximum duration of 10 seconds. 
We uniformly sample 10 frames per video and perform object detection using DETR~\cite{carion2020end}. 
Detected objects are used to construct the final ground-truth object sets for hallucination evaluation.

\subsection{Adaptive Sink-Guided Decoding Algorithm}
\label{subsec:algo}
\subsubsection{Insignificance of Object Tokens in Hallucination Detection}
\label{just}
\begin{figure}[htbp]
    \centering
    \includegraphics[width=0.5\linewidth]{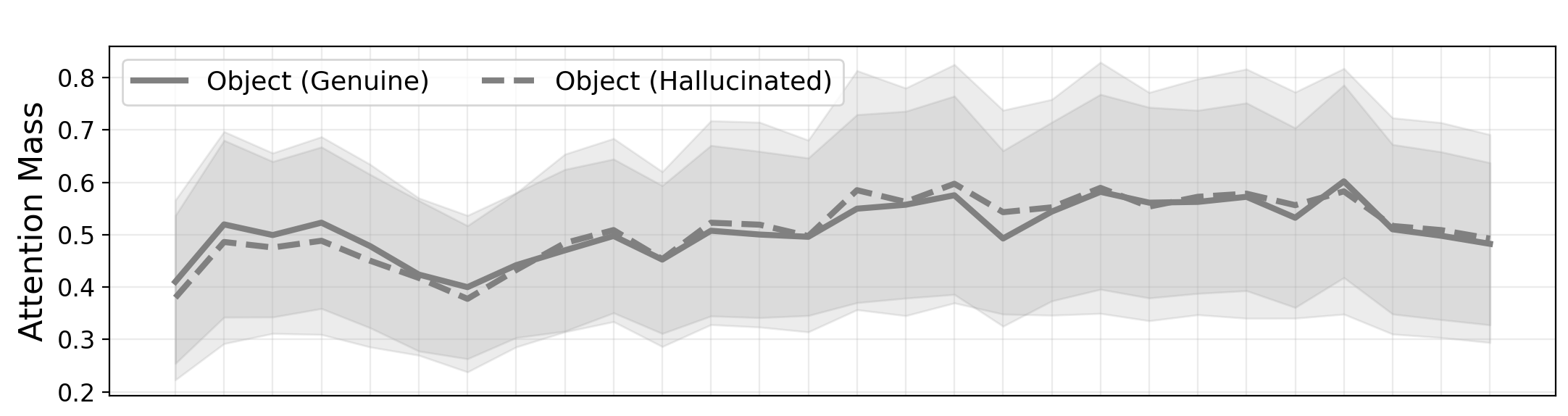}
    \caption{Comparison of attention weights on object tokens during the generation of genuine versus hallucinated objects. Unlike the unimodal sink, object tokens do not show a significant disparity between the two scenarios.}
    \label{fig:object_attn}
\end{figure}

While \cref{hall_attn} demonstrated the sensitivity of sink tokens to hallucination, we investigate if object tokens share this characteristic. Contrary to sink tokens, \cref{fig:object_attn} reveals that attention on object tokens shows no significant divergence between genuine and hallucinated cases. This invariance sharply contrasts with the distinct attention disparity observed in sink tokens (\cref{fig:cross_uni}).
   
\subsubsection{Details of Stability Functions}
\label{appendix:smoothing}

To ensure stability and prevent abrupt shifts caused by fluctuations in the raw coefficient, we apply threshold-based gating and momentum-based temporal smoothing to $\gamma_t^{\text{base}}$ to obtain the stabilized coefficient $\gamma_t$.

To compute $\gamma_t$, we first define the base adaptive weight $\gamma_t^{\text{base}}$, which quantifies the relative dominance of unimodal versus cross-modal sink attention at decoding step $t$:
\begin{equation}
\gamma_t^{\text{base}}
=
\frac{
\bar{A}_{t,\mathrm{uni}}
}{
\bar{A}_{t,\mathrm{uni}} + \bar{A}_{t,\mathrm{cross}} + \epsilon
},
\end{equation}
where $\bar{A}_{t,\mathrm{uni}}$ and $\bar{A}_{t,\mathrm{cross}}$ denote the average attention mass assigned to unimodal and cross-modal sink tokens, respectively, and $\epsilon$ is a small constant for numerical stability.

Concretely, the target coefficient is defined as:
\begin{equation}
\hat{\gamma}_t =
\begin{cases}
0, & \gamma_t^{\text{base}} < \tau \ \text{or}\ r_t > \rho, \\[4pt]
\gamma_{\max} \gamma_t^{\text{base}}, & \text{otherwise},
\end{cases}
\label{eq:gamma_target_piecewise}
\end{equation}
where $\gamma_{\max}$ is set to $0.6$, $\tau$ is the adaptive gating threshold set to $0.6$, $r_t$ denotes the average attention mass assigned to text tokens, and $\rho$ is the text-mass threshold set to $0.5$.

Finally, we apply momentum-based temporal smoothing:
\begin{equation}
\gamma_t
=
\beta \gamma_{t-1}
+
(1-\beta)\hat{\gamma}_t,
\label{eq:gamma_momentum}
\end{equation}
where $\beta$ is the momentum coefficient set to $0.7$, which prevents abrupt temporal fluctuations during decoding.

\subsection{Details of Adapting Baseline Approaches to Audio-Visual Setting}
\label{appendix:baseline_details}

In this section, we describe how we adapted existing image-based baselines for the audio-visual setting.

\textbf{PAI}
PAI~\cite{liu2024payingattentionimagetrainingfree} aims to mitigate hallucinations by amplifying the attention scores of image tokens globally. The original mechanism is formulated as:
\begin{align}
\tilde{A}_{t,j} &\leftarrow {A}_{t,j} + \alpha \lvert {A}_{t,j} \rvert, \quad j \in \mathcal{S}_{\text{image}},
\end{align}
where $\mathcal{S}_{\text{image}}$ represents the set of image indices. To adapt PAI for our audio-visual setting, we modify the method to increase the attention weights for both audio and video tokens, rather than image tokens alone.

\textbf{VCD}
VCD~\cite{leng2023mitigatingobjecthallucinationslarge} reduces statistical priors and hallucinations by contrasting the original output logits with logits derived from distorted visual inputs. While the original VCD applies noise solely to the image modality, we extend this approach to the audio-visual domain. Specifically, we apply noise to both audio and video inputs to generate the distorted logits for contrastive decoding.

\section{Additional Analysis}
\label{sec:add_analysis}
\subsection{Causal Tracing Analysis}
\label{appendix:causal_tracing_further}

\subsubsection{Bottom-up Analysis}
\label{appendix:bottom_up}
To further verify whether our pre-selected token categories indeed coincide with the tokens that are causally most relevant to cross-modal information flow, we additionally conduct a bottom-up, token-level causal analysis on 100 samples using Qwen2.5-Omni(7B).
Specifically, we repeat the experiments in \cref{exp} by applying causal patching to individual tokens rather than predefined groups.
After ranking the tokens by their causal effect $\Delta$, we measure the proportion of sink and object tokens ($\mathcal{S}\cup\mathcal{O}$) versus the remaining (\emph{Neither}) tokens among the top $5\%$, $10\%$, and $20\%$ most influential tokens. As shown in \cref{tab:bottom_up_topk}, the majority of the most influential tokens fall into $\mathcal{S}\cup\mathcal{O}$.
This supports our premise that sink and object tokens are highly meaningful, providing a reasonable basis for our initial hypothesis.

\begin{table}[htbp]
\caption{Proportion of sink/object tokens ($\mathcal{S}\cup\mathcal{O}$) versus the remaining (\emph{Neither}) tokens among the top-$k\%$ most influential tokens, ranked by their causal effect $\Delta$. $N$ denotes the parameter for defining global sink tokens (\cref{exp}).}
\label{tab:bottom_up_topk}
\centering
\setlength{\tabcolsep}{6pt}
\renewcommand{\arraystretch}{1.15}
\begin{tabular}{cl ccc ccc}
\toprule
& & \multicolumn{3}{c}{\textbf{Video Dom.}} & \multicolumn{3}{c}{\textbf{Audio Dom.}} \\
\cmidrule(lr){3-5} \cmidrule(lr){6-8}
$N$ & Group & $5\%$ & $10\%$ & $20\%$ & $5\%$ & $10\%$ & $20\%$ \\
\midrule
\multirow{2}{*}{2}
    & $\mathcal{S}\cup\mathcal{O}$ & 0.88 & 0.90 & 0.84 & 0.71 & 0.73 & 0.77 \\
    & Neither                       & 0.12 & 0.10 & 0.16 & 0.29 & 0.27 & 0.23 \\
\midrule
\multirow{2}{*}{3}
    & $\mathcal{S}\cup\mathcal{O}$ & 0.82 & 0.84 & 0.84 & 0.64 & 0.66 & 0.70 \\
    & Neither                       & 0.18 & 0.16 & 0.16 & 0.36 & 0.34 & 0.30 \\
\bottomrule
\end{tabular}
\end{table}

To further quantify how much cross-modal information resides outside $\mathcal{S}\cup\mathcal{O}$, we measure the proportion (\%) of the causal effect $\Delta$ captured by patching the $\mathcal{S}\cup\mathcal{O}$ group versus all remaining tokens (\emph{Neither}), relative to the effect of patching all tokens. Our results in \cref{tab:bottom_up_so_vs_neither} indicate that while some cross-modal information may be distributed outside these key tokens, this fraction is minimal. We clarify that we do not claim cross-modal information is \emph{exclusively} stored within our identified hubs; rather, they serve as its disproportionately dominant hubs.
\begin{table}[htbp]
\caption{Proportion of the causal effect $\Delta$ captured by the $\mathcal{S}\cup\mathcal{O}$ group versus the \emph{Neither} group, relative to patching all tokens.}
\label{tab:bottom_up_so_vs_neither}
\centering
\setlength{\tabcolsep}{10pt}
\renewcommand{\arraystretch}{1.15}
\begin{tabular}{cl cc}
\toprule
$N$ & Group & \textbf{Video Dom.} & \textbf{Audio Dom.} \\
\midrule
\multirow{2}{*}{2}
    & $\mathcal{S}\cup\mathcal{O}$ & 0.92 & 0.83 \\
    & Neither                       & 0.11 & 0.20 \\
\midrule
\multirow{2}{*}{3}
    & $\mathcal{S}\cup\mathcal{O}$ & 0.87 & 0.73 \\
    & Neither                       & 0.16 & 0.29 \\
\bottomrule
\end{tabular}
\end{table}

\subsubsection{Patching Locations Analysis}
\label{appendix:patching_location}

Identifying the optimal layer for hidden state patching is a critical component of the causal tracing framework. Prior studies investigating factual association in text-only LLMs and Large Vision-Language Models (LVLMs) typically compare the efficacy of patching immediately after the Self-Attention (SA) mechanism versus after the MLP layers \cite{meng2023locatingeditingfactualassociations}. These works generally conclude that factual knowledge is predominantly stored within the MLP layers.

However, our research objective differs significantly. Rather than locating static factual storage, we aim to trace the dynamic flow of cross-modal information between audio and video inputs. Given that cross-modal fusion relies on the token-mixing capabilities of the attention mechanism, we posit that the Self-Attention layers are the primary locus of this interaction.

Furthermore, we critically assess whether to patch \emph{before} or \emph{after} the Self-Attention (SA) operation. While the output states \emph{after} the SA layer represent the result of information fusion, the critical limitation is that the actual transfer of this fused information to the text modality occurs \emph{during} the attention computation. Consequently, if we patch the output of the SA layer, the fused cross-modal information is not effectively attended to by the text tokens within that layer, resulting in a failure to reflect this information in the model's final prediction. 

Therefore, we hypothesize that patching the hidden states \emph{before} the SA layer (i.e., the input to the SA block) is methodologically superior. Although the pre-attention states may theoretically contain less fully integrated cross-modal information compared to the post-attention outputs, patching at this stage is crucial as it allows the attention mechanism to actively propagate the restored information to the text sequence, thereby ensuring it is incorporated into the model's output.

To validate this hypothesis, we conducted a control experiment distinct from our main analysis. Unlike the main experiments, which patch the hidden states of the non-dominant modality (where the dominant information is only partially integrated, preventing the prediction probability from being fully restored to the clean run level), here we patched the \emph{dominant} modality. In this setting, since the dominant modality dictates the correct answer, successfully patching its information should result in a near-complete recovery of the correct prediction (i.e., a high $\mathrm{IE}_\text{clean}$ and $\mathrm{IE}_\text{corrupt}$).

We compared patching performance at three specific locations: (1) Before Self-Attention, (2) After Self-Attention, and (3) After MLP. We evaluated the restoration of correct answers using the Qwen2.5-Omni (7B) model in Audio-Dominant scenarios. The results, presented in \cref{tab:patching_location_comparison}, demonstrate that patching \textbf{Before Self-Attention} yields a significantly higher recovery of the correct prediction compared to other locations. Consequently, we decided to patch the hidden states before the self-attention layer for all main experiments.

\begin{table}[h]
    \centering
    \caption{Comparison of Indirect Effects (IE) across different patching locations on Qwen2.5-Omni (7B) under Audio-Dominant settings. Patching all dominant tokens before the Self-Attention layer results in the most effective information recovery.}
    \label{tab:patching_location_comparison}
    \renewcommand{\arraystretch}{1.2}
    \begin{tabular}{lcc}
        \toprule
        \textbf{Patching Location} & \textbf{IE\textsubscript{clean}} & \textbf{IE\textsubscript{corr}} \\
        \midrule
        Before Self-Attention & 63.72 & 35.22 \\
        After Self-Attention  & 3.01 & 2.38 \\
        After MLP Layer       & 20.96 & 11.67 \\
        \bottomrule
    \end{tabular}
\end{table}
\subsubsection{Causal Tracing Results with Different Patching Locations}

\label{app:causl_tracing_ablation} As discussed earlier, although we primarily patch hidden states at the self-attention module, which we find to be the most suitable patching location, we further examine the robustness of our findings with respect to the patching location. Specifically, following prior work \cite{meng2023locatingeditingfactualassociations,basu2024understandinginformationstoragetransfer,li2025causaltracingobjectrepresentations}, we consider two alternative patching points:
(i) immediately after the self-attention block, and
(ii) immediately after the MLP block.

The causal tracing results obtained by patching hidden states after the self-attention block are reported in \cref{tab:self_attn_ablation}, while those obtained by patching hidden states after the MLP block are shown in \cref{tab:mlp_ablation}.
Consistent with the main results, we observe that cross-modal information is predominantly concentrated in sink tokens, with cross-modal sink tokens exhibiting the strongest effects.

\input{table/ablation_causal_tracing1}

\input{table/ablation_causal_tracing2}

\subsubsection{Layerwise Analysis}
\label{subsubsec:layerwise}
\begin{figure}[htbp]
    \centering
    \includegraphics[width=0.4\linewidth]{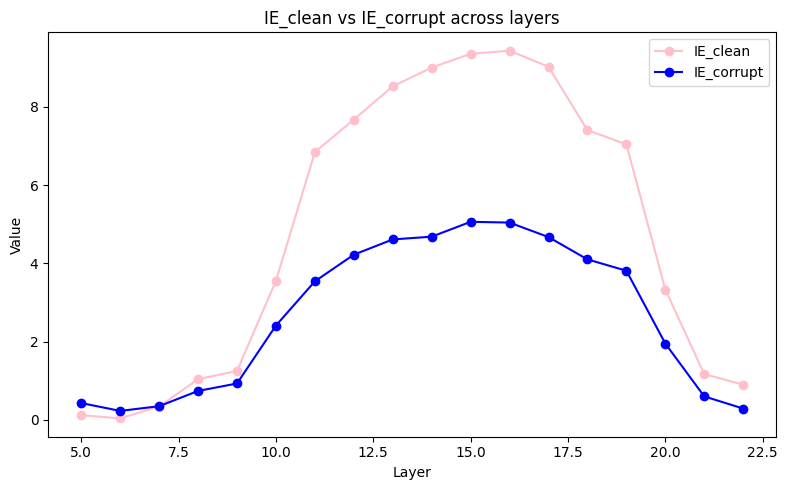}
    \caption{
    Layer-wise causal patching results for the audio-dominant setting on Qwen2.5-Omni (7B).
    We patch all tokens from the non-dominant modality using a sliding window of 10 consecutive layers.
    cross-modal information exchange is most pronounced in the middle layers, indicating that these layers act as primary integration hubs.
    }
    \label{fig:output}
\end{figure}

\cref{fig:output} shows layer-wise causal patching results for the audio-dominant setting on Qwen2.5-Omni (7B), where we patch all tokens from the non-dominant modality.
We use a sliding window of size 10, patching tokens across 10 consecutive layers at a time.
The results indicate that cross-modal information exchange is most pronounced in the middle layers.

\subsubsection{Alternative Corruptions Method Analysis}
\label{app:corruption_alt}
To verify that our results are robust to corruption choices beyond zeroing out, we repeat the experiments in \cref{exp} on Qwen2.5-Omni(7B) using two alternative corruption methods: (1)~injecting Gaussian noise into the raw input, and (2)~replacing the multimodal encoder output embeddings with their mean embedding, following~\cite{neotowards}. As shown in \cref{tab:corruption_alt}, both alternatives align with our findings.

\begin{table}[htbp]
\caption{Causal tracing results on Qwen2.5-Omni(7B) under two alternative corruption methods: (1)~injecting Gaussian noise into the raw input, and (2)~replacing the encoder output embeddings with their mean.}
\label{tab:corruption_alt}
\centering
\small
\setlength{\tabcolsep}{6pt}
\renewcommand{\arraystretch}{1.15}
\begin{tabular}{ll cc cc}
\toprule
& & \multicolumn{2}{c}{\textbf{(1)~Noise}} & \multicolumn{2}{c}{\textbf{(2)~Encoder}} \\
\cmidrule(lr){3-4} \cmidrule(lr){5-6}
\textbf{Modality} & \textbf{Ablation}
& $\mathrm{IE}_{\text{clean}}\uparrow$ & $\mathrm{IE}_{\text{corr}}\uparrow$
& $\mathrm{IE}_{\text{clean}}\uparrow$ & $\mathrm{IE}_{\text{corr}}\uparrow$ \\
\midrule
\multirow{10}{*}{\textbf{Audio Dominant}}
& All              & 9.6 & 5.3 & 6.7 & \phantom{$-$}2.5 \\
\cmidrule(lr){2-6}
& Object           & 5.1 & 2.4 & 4.1 & \phantom{$-$}0.5 \\
\cmidrule(lr){2-6}
& Random ($N\!=\!2$) & 4.5 & 2.2 & 3.3 & $-0.3$ \\
& Random ($N\!=\!3$) & 2.7 & 1.8 & 1.9 & $-1.5$ \\
\cmidrule(lr){2-6}
& Sink ($N\!=\!2$)       & 6.2 & 2.9 & 3.6 & $-0.2$ \\
& Sink ($N\!=\!3$)       & 4.3 & 2.0 & 2.6 & $-0.9$ \\
\cmidrule(lr){2-6}
& Unimodal ($N\!=\!2$)   & 0.7 & 0.2 & 0.6 & $-2.2$ \\
& Unimodal ($N\!=\!3$)   & 0.9 & 0.4 & 0.5 & $-2.5$ \\
\cmidrule(lr){2-6}
& Crossmodal ($N\!=\!2$) & 5.6 & 3.0 & 3.2 & $-0.5$ \\
& Crossmodal ($N\!=\!3$) & 3.6 & 1.5 & 2.0 & $-1.6$ \\
\midrule
\multirow{10}{*}{\textbf{Video Dominant}}
& All              & 8.4 & 13.5 & 12.8 & 13.5 \\
\cmidrule(lr){2-6}
& Object           & 5.4 & \phantom{0}8.4  & \phantom{0}7.2  & \phantom{0}6.6 \\
\cmidrule(lr){2-6}
& Random ($N\!=\!2$) & 4.5 & \phantom{0}6.4 & \phantom{0}7.8 & \phantom{0}7.0 \\
& Random ($N\!=\!3$) & 2.8 & \phantom{0}3.7 & \phantom{0}4.4 & \phantom{0}4.0 \\
\cmidrule(lr){2-6}
& Sink ($N\!=\!2$)       & 5.8 & \phantom{0}8.5 & \phantom{0}8.7 & \phantom{0}8.9 \\
& Sink ($N\!=\!3$)       & 4.7 & \phantom{0}7.0 & \phantom{0}7.2 & \phantom{0}6.9 \\
\cmidrule(lr){2-6}
& Unimodal ($N\!=\!2$)   & 2.1 & \phantom{0}3.4 & \phantom{0}3.8 & \phantom{0}3.5 \\
& Unimodal ($N\!=\!3$)   & 1.9 & \phantom{0}3.0 & \phantom{0}3.2 & \phantom{0}2.4 \\
\cmidrule(lr){2-6}
& Crossmodal ($N\!=\!2$) & 3.3 & \phantom{0}4.4 & \phantom{0}5.2 & \phantom{0}4.3 \\
& Crossmodal ($N\!=\!3$) & 2.4 & \phantom{0}3.7 & \phantom{0}4.1 & \phantom{0}3.1 \\
\bottomrule
\end{tabular}
\end{table}

\subsubsection{MDS Statistics}
\label{app:mds_stat}
We analyze MDS statistics of video and audio sink tokens for 100 samples (\cref{tab:mds_stats}). Video and audio sink tokens exhibit positive and negative median values, respectively, indicating that sink tokens tend to receive more attention from their own modality than from complementary modality. However, the distribution is far from degenerate-given that MDS is bounded within [-1,1], the sizable IQR and Std show considerable variation in the proportion of attention each sink token receives from two modalities, supporting \cref{sec:dissec} and \cref{fig:mds_visualize}.
\begin{table}[htbp]
\caption{Distribution statistics of the Modality Dominance Score (MDS) for video and audio sink tokens, computed over 100 samples across five backbones. }
\label{tab:mds_stats}
\centering
\setlength{\tabcolsep}{6pt}
\renewcommand{\arraystretch}{1.15}
\resizebox{\textwidth}{!}{%
\begin{tabular}{ll ccccc}
\toprule
\textbf{Metric} & \textbf{Modality}
& \textbf{Qwen2.5-Omni(7B)} & \textbf{Qwen2.5-Omni(3B)}
& \textbf{video-SALMONN-o1(7B)} & \textbf{video-SALMONN2+(7B)} & \textbf{video-SALMONN2+(3B)} \\
\midrule
\multirow{2}{*}{Median}
    & Video & \phantom{-}0.45 & \phantom{-}0.49 & \phantom{-}0.56 & \phantom{-}0.84 & \phantom{-}0.79 \\
    & Audio & $-0.49$         & $-0.50$         & $-0.77$         & $-0.18$         & $-0.59$         \\
\midrule
\multirow{2}{*}{IQR}
    & Video & 0.34 & 0.34 & 0.39 & 0.13 & 0.22 \\
    & Audio & 0.23 & 0.27 & 0.18 & 0.73 & 0.77 \\
\midrule
\multirow{2}{*}{Std}
    & Video & 0.25 & 0.26 & 0.30 & 0.14 & 0.22 \\
    & Audio & 0.26 & 0.27 & 0.16 & 0.55 & 0.46 \\
\bottomrule
\end{tabular}%
}
\end{table}

\subsection{Object Hallucination Reduction}
\label{app:obj_hall_red}
\subsubsection{Object Hallucination Case Analyses}
\label{addition:object_hall_case}
\cref{tab:hall_cases} quantifies the prevalence of such cross-modal disagreement cases and the hallucinated cases with audiovisual disagreement cases.

\begin{table}[htbp]
\centering
\caption{Hallucination statistics on VGGSound-Animal dataset.}
\label{tab:hall_cases}
\resizebox{0.5\columnwidth}{!}{
\begin{tabular}{l c}
\toprule
\textbf{Metric} &\textbf{VGGSound-Animal} \\
\midrule
Total samples & 363 \\
\midrule
Total audio-visual disagreement samples & 272\\
Total hallucinated samples & 175 \\
Hallucinated samples with A--V disagreement & 72 \\
\bottomrule
\end{tabular}
}
\end{table}

\subsubsection{Additional Results on Different Models}
\label{app:add_models}

We extend our evaluation to verify whether ASD effectively mitigates hallucinations at the 3B-scale model.
As shown in \cref{tab:asd_3b}, applying ASD ($\alpha\!=\!0.3,\ N=\!2$) to Qwen2.5-Omni(3B) consistently reduces hallucination metrics across all three datasets, demonstrating the generalizability of our method.

We also note the distinct behavior of video-SALMONN2+.
Unlike Qwen2.5-Omni and video-SALMONN-o1, the training paradigm of video-SALMONN2+ incorporates explicit mechanisms to suppress caption hallucinations.
As a result, its baseline hallucination scores are already exceptionally low-showing an average $\mathrm{C}_s$ reduction of $43.9\%$ and $21.1\%$ across the three datasets for the 7B and 3B variants, respectively, compared to video-SALMONN-o1(7B) leaving a narrower margin for further improvement via training-free interventions.

Unlike approaches that require significant training overhead, ASD provides a \textbf{plug-and-play, training-free} alternative.
Furthermore, we emphasize that our core contribution lies in the identification of \emph{cross-modal information hubs}, which we leverage to effectively reduce captioning hallucinations across diverse models with varying scales and architectures.

\begin{table}[htbp]
\caption{Quantitative evaluation of ASD applied to Qwen2.5-Omni(3B) using CHAIR for hallucination, and F1 scores for caption richness.}
\label{tab:asd_3b}
\centering
\setlength{\tabcolsep}{6pt}
\renewcommand{\arraystretch}{1.15}
\begin{tabular}{llccc}
\toprule
\textbf{Dataset} & \textbf{Method} & $\mathrm{C}_s \downarrow$ & $\mathrm{C}_i \downarrow$ & F1 $\uparrow$ \\
\midrule
\multirow{2}{*}{VGGSound-Animal}
    & Vanilla       & 51.24 & 41.00 & 52.88 \\
    & \textbf{ASD}  & \textbf{43.25} & \textbf{38.52} & \textbf{50.44} \\
\midrule
\multirow{2}{*}{VGGSound-All}
    & Vanilla       & 40.52 & 27.15 & 57.86 \\
    & \textbf{ASD}  & \textbf{37.25} & \textbf{26.01} & \textbf{57.59} \\
\midrule
\multirow{2}{*}{Audioset}
    & Vanilla       & 16.23 & 15.79 & 70.80 \\
    & \textbf{ASD}  & \textbf{15.50} & \textbf{15.34} & \textbf{70.28} \\
\bottomrule
\end{tabular}
\end{table}

\subsubsection{Reverse ASD as a Counterfactual Test}
\label{app:reverse_asd}
To further validate that the gains of ASD indeed stem from selectively promoting cross-modal sink tokens, we conduct a counterfactual experiment using a \emph{reverse} variant of our method.
Concretely, we reverse the signs in Eq.~(6) and Eq.~(7), so that the intervention decreases attention to cross-modal sink tokens while correspondingly increasing attention to unimodal sink tokens.
We additionally scale the base guidance coefficient $\alpha$ proportionally to the total sink attention attributed to cross-modal sinks. This design ensures that when attention to cross-modal sinks is naturally high, the reverse ASD mechanism actively penalizes it while boosting attention to unimodal sinks.

\begin{table}[htbp]
\caption{Effect of \emph{reverse} ASD on Qwen2.5-Omni(7B) across three datasets. }
\label{tab:reverse_asd}
\centering
\setlength{\tabcolsep}{10pt}
\renewcommand{\arraystretch}{1.15}
\begin{tabular}{ll ccc}
\toprule
\textbf{Dataset} & \textbf{Method}
& $\mathrm{C}_s \downarrow$ & $\mathrm{C}_i \downarrow$ & F1 $\uparrow$ \\
\midrule
\multirow{2}{*}{VGGSound-Animal}
    & Vanilla       & 48.21 & 37.13 & 55.24 \\
    & Reverse ASD   & 46.83 & 41.74 & 49.14 \\
\midrule
\multirow{2}{*}{VGGSound-All}
    & Vanilla       & 30.70 & 20.67 & 58.69 \\
    & Reverse ASD   & 33.39 & 24.50 & 54.31 \\
\midrule
\multirow{2}{*}{Audioset}
    & Vanilla       & \phantom{0}8.92 & 10.93 & 69.73 \\
    & Reverse ASD   & \phantom{0}9.65 & 10.59 & 71.32 \\
\bottomrule
\end{tabular}
\end{table}
\cref{tab:reverse_asd} presents the results of applying this reverse ASD on Qwen2.5-Omni(7B) across three datasets. As hypothesized, the overall performance deteriorates. However, we observe that the degradation is not perfectly symmetric to the improvements gained by the original ASD, consistent with the analysis in \cite{asd2025}. In hallucination-unprone states, the decoding process is firmly supported by stable evidence. Consequently, artificially increasing attention to unimodal sinks in these stable states does not easily induce new hallucinations. Furthermore, in already hallucination-prone states, where hallucinations are already occurring, further amplifying unimodal sink attention does not significantly induce additional hallucinations.

\subsubsection{Comparison with inference-time interventions in AVLLMs}
\label{app:av_decoding_baselines}
We further compare ASD against two recent decoding methods designed for AVLLMs: AVCD~\cite{avcd} and FMD~\cite{fmd}.
As shown in \cref{tab:decoding_baselines}, ASD consistently achieves the strongest overall performance across the three datasets, attaining the best $\mathrm{C}_s$ and $\mathrm{C}_i$ scores on every benchmark and the highest ALOHa on the two VGGSound subsets.

\begin{table}[htbp]
\caption{Comparison with decoding-based hallucination mitigation baselines for AVLLMs on Qwen2.5-Omni(7B). We report ALOHa and CHAIR metrics for hallucination, and F1 for caption richness. \textbf{Bold} and \underline{underlined} values indicate the best and second-best results within each dataset.}
\label{tab:decoding_baselines}
\centering
\setlength{\tabcolsep}{7pt}
\renewcommand{\arraystretch}{1.15}
\begin{tabular}{ll cccc}
\toprule
\textbf{Dataset} & \textbf{Method}
& ALOHa $\uparrow$ & $\mathrm{C}_s \downarrow$ & $\mathrm{C}_i \downarrow$ & F1 $\uparrow$ \\
\midrule
\multirow{4}{*}{VGGSound-Animal}
    & Vanilla & \underline{40.7} & \underline{48.2} & \underline{37.1} & \textbf{55.2} \\
    & AVCD    & 38.5             & 51.5             & 38.8             & \underline{53.7} \\
    & FMD     & 37.3             & 50.9             & 40.2             & 52.9 \\
    & ASD     & \textbf{42.7}    & \textbf{36.9}    & \textbf{34.1}    & 52.4 \\
\midrule
\multirow{4}{*}{VGGSound-All}
    & Vanilla & \underline{35.0} & \underline{30.7} & \textbf{20.6}    & \textbf{58.6} \\
    & AVCD    & 32.8             & 33.5             & 24.4             & 55.7 \\
    & FMD     & 32.5             & 37.9             & 25.9             & \underline{57.3} \\
    & ASD     & \textbf{38.8}    & \textbf{29.6}    & \underline{21.7} & 55.8 \\
\midrule
\multirow{4}{*}{Audioset}
    & Vanilla & 38.2             & \underline{\phantom{0}8.9} & \underline{10.9} & 69.7 \\
    & AVCD    & \underline{59.3} & 10.8                       & 12.1             & 69.2 \\
    & FMD     & \textbf{63.7}    & 13.6                       & 13.5             & \underline{71.2} \\
    & ASD     & 38.2             & \textbf{\phantom{0}8.5}    & \textbf{10.2}    & \textbf{72.9} \\
\bottomrule
\end{tabular}
\end{table}
\subsubsection{Latency Overhead of ASD}
\label{app:overhead}
Although ASD proves effective, it comes at the cost of increased inference latency (3.7× slower ms/token), arising from the sink identification step and the dual forward passes it requires.

\section{Qualitative Results}
\label{app:qualitative_results}
\cref{fig:qualitative_results} shows qualitative results of our method. The baseline produces hallucinated object descriptions, whereas our method includes only genuine objects.
\begin{figure}[t]
    \centering
    \includegraphics[width=0.85\linewidth]{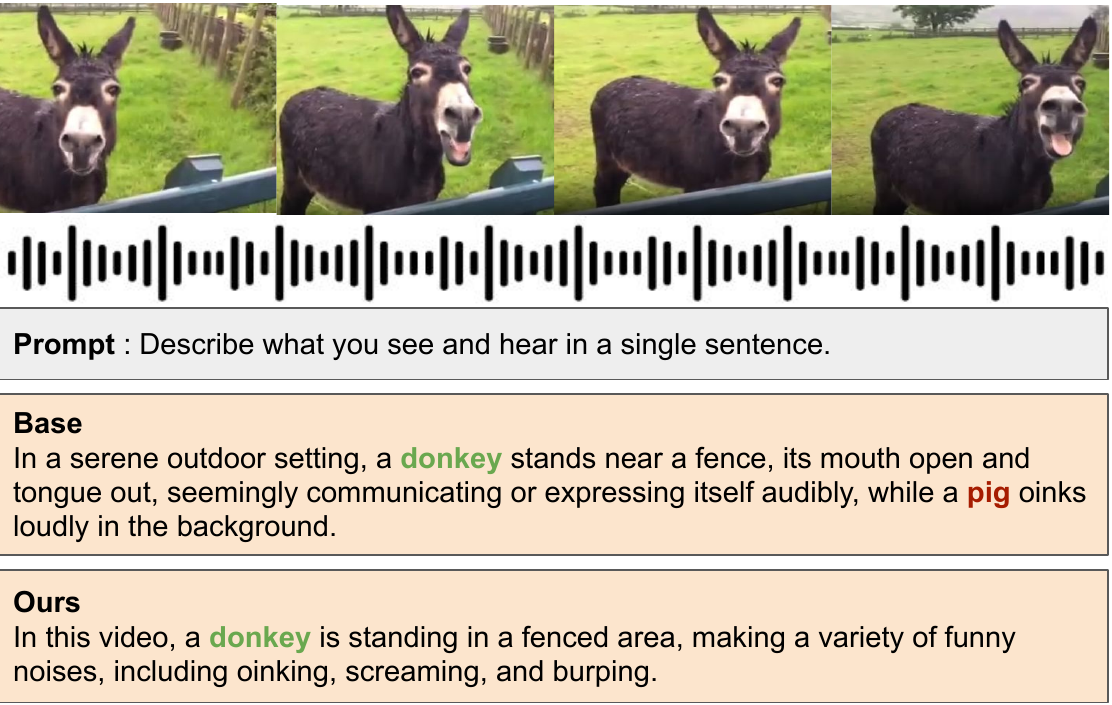}
    \vspace{0.7em}

    \includegraphics[width=0.85\linewidth]{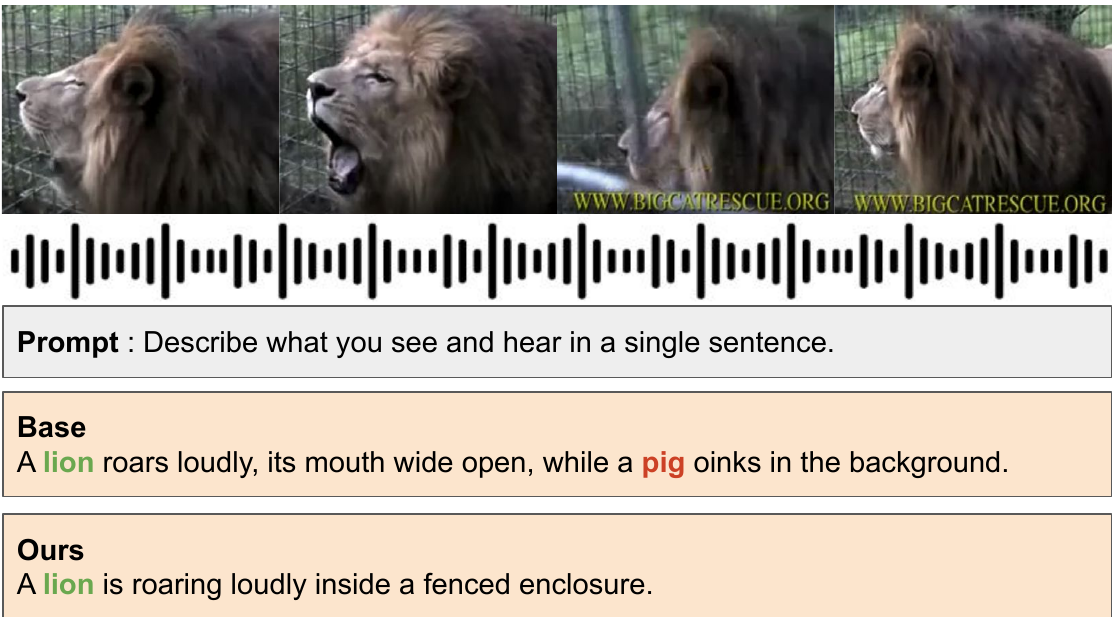}

    \caption{Qualitative results of our method. 
Text highlighted in red indicates hallucinated objects, while unhighlighted text denotes genuine objects.}
    \label{fig:qualitative_results}
\end{figure}

\begin{figure}[t]
    \ContinuedFloat
    \centering

    \includegraphics[width=0.85\linewidth]{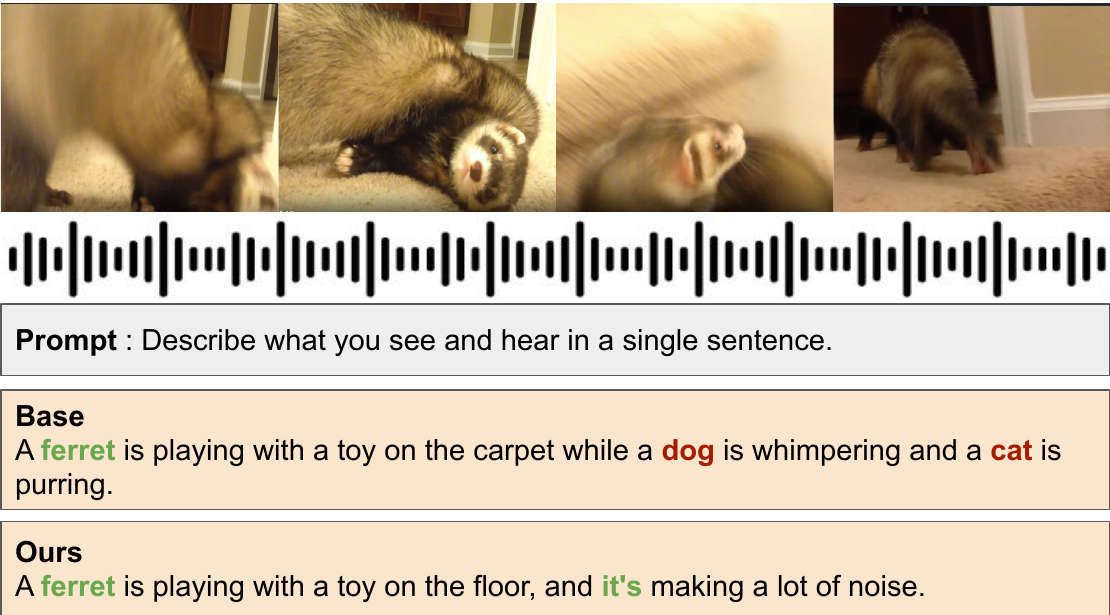}

    \vspace{0.7em}

    \includegraphics[width=0.85\linewidth]{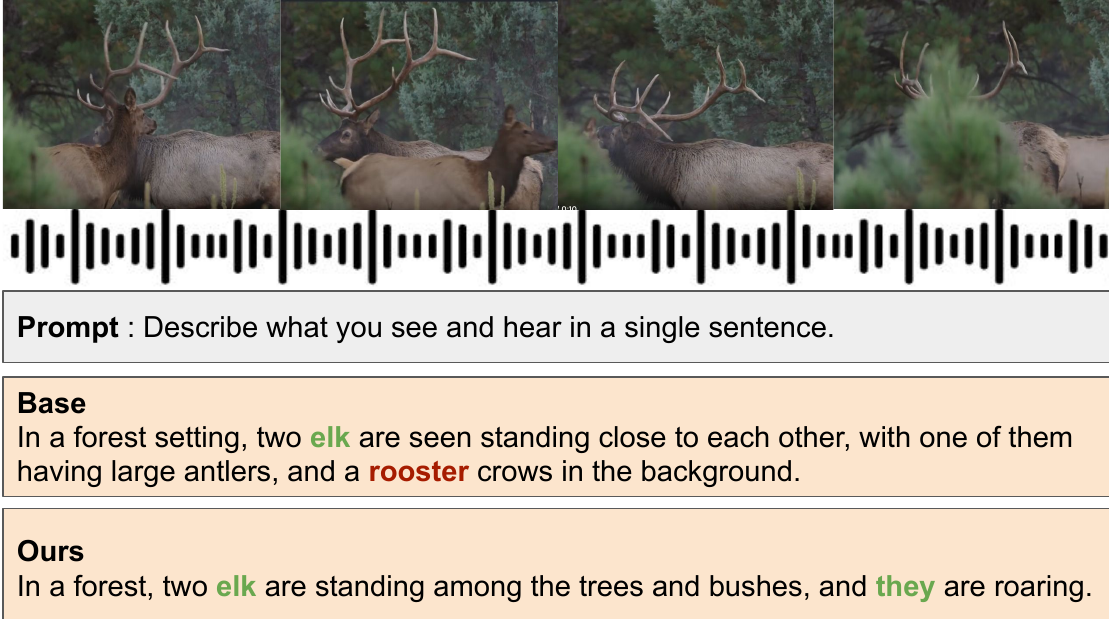}

    \caption{Qualitative results of our method. Continued. Text highlighted in red indicates hallucinated objects, while unhighlighted text denotes genuine objects.}
    \label{fig:qualitative_results_continued}
\end{figure}

\section{Limitations}
\label{app:limit}
Our work has several limitations. First, due to memory constraints, we were unable to evaluate our analysis on larger models such as Qwen3-Omni (30B)\cite{xu2025qwen3omnitechnicalreport}, leaving the 
scalability of our findings to such scales as an open question. Second, our proposed ASD method is applicable only to captioning tasks and does not extend to general question answering. Finally, ASD introduces a non-trivial latency overhead, which may limit its practicality in latency-sensitive applications.

%% file: table/appendix_dataset_stats.tex
\begin{table}[htbp]
\centering
\scriptsize
\caption{Number of samples retained for causal tracing analysis after applying
audio-dominant and video-dominant filtering criteria for each model.}
\label{tab:appendix_dataset_stats}
\setlength{\tabcolsep}{3pt}
\begin{tabular}{lccccc}
\toprule
\textbf{Dominance} &
\textbf{Qwen2.5-Omni (7B)} &
\textbf{Qwen2.5-Omni (3B)} &
\textbf{video-SALMONN-o1 (7B)} &
\textbf{video-SALMONN2+ (7B)} &
\textbf{video-SALMONN2+ (3B)} \\
\midrule
Audio Dominant & 349 & 315 & 184 & 369 & 141 \\
Video Dominant & 548 & 646 & 828 & 808 & 890 \\
\bottomrule
\end{tabular}
\end{table}

%% file: table/ablation_causal_tracing1.tex
\begin{table*}[htbp]
\centering
\tiny
\fontsize{5.5pt}{6.5pt}\selectfont
\caption{Causal tracing results obtained by patching hidden states after the self-attention block.
}
\label{tab:self_attn_ablation}
\scriptsize
\setlength{\tabcolsep}{3pt}

\newcolumntype{C}[1]{>{\centering\arraybackslash}p{#1}}
\resizebox{\textwidth}{!}
{
    \begin{tabular}{ll|
    C{0.85cm}C{0.85cm}>{\columncolor{gray!15}}C{0.7cm}|
    C{0.85cm}C{0.85cm}>{\columncolor{gray!15}}C{0.7cm}|
    C{0.85cm}C{0.85cm}>{\columncolor{gray!15}}C{0.7cm}|  
    C{0.85cm}C{0.85cm}>{\columncolor{gray!15}}C{0.7cm}|
    C{0.85cm}C{0.85cm}>{\columncolor{gray!15}}C{0.7cm}
    }

    \toprule
    \textbf{Modality} & \textbf{Ablation}
    & \multicolumn{3}{c|}{\textbf{Qwen2.5-Omni (7B)}}
    & \multicolumn{3}{c|}{\textbf{Qwen2.5-Omni (3B)}}
    & \multicolumn{3}{c|}{\textbf{video-SALMONN-o1 (7B)}} 
    & \multicolumn{3}{c|}{\textbf{video-SALMONN2+ (7B)}}
    & \multicolumn{3}{c}{\textbf{video-SALMONN2+ (3B)}} \\
    
    \cmidrule(lr){3-5}
    \cmidrule(lr){6-8}
    \cmidrule(lr){9-11}   
    \cmidrule(lr){12-14}  
    \cmidrule(lr){15-17}  

    &
    & IE$_{\text{clean}}$ $\uparrow$ & IE$_{\text{corr}}$ $\uparrow$ & \#Tokens
    & IE$_{\text{clean}}$ $\uparrow$ & IE$_{\text{corr}}$ $\uparrow$ & \#Tokens
    & IE$_{\text{clean}}$ $\uparrow$ & IE$_{\text{corr}}$ $\uparrow$ & \#Tokens  
    & IE$_{\text{clean}}$ $\uparrow$ & IE$_{\text{corr}}$ $\uparrow$ & \#Tokens
    & IE$_{\text{clean}}$ $\uparrow$ & IE$_{\text{corr}}$ $\uparrow$ & \#Tokens \\
    
    \midrule
    
    \multirow{14}{*}{\makecell{\textbf{Audio}\\\textbf{Dominant}}}
    & All
    & 9.61 &    5.29  &1440
    &   7.83	& 3.48& 1440
    & 35.55	&	33.18	&	1820
    &    6.45	&	5.27	&	1210
    &    1.92	&	2.15	&	1210  \\
    \addlinespace[0.2em]
    \cmidrule(lr){2-17}

    & Object
    &   5.05 &	2.44     &613
    &    3.53	&	1.12	&	580
    &   16.22	&	15.06	&	852
    &   3.78	&	3.93	&	500
    &     0.72	&	1.16	&	448  \\
    \cmidrule(lr){2-17}

    & Random (N=2)
    &     4.01	&2.34	&603
    &      4.02 &	1.19 &	605
    &    19.19	&	17.98	&	818
    &     4.27	&	4.09	&	566
    &     0.94	&	1.12	&	506   \\
    & Random (N=3)
    &  2.76	 &    1.14 &    	363
    &     2.21&	0.50&	355
    &     13.01	&	12.51	&	514
    &     3.17	&	3.53	&	360
    &     0.83	&	0.77	&	298  \\
    
    & Random (N=4)
    &     1.93	 & 1.02 & 	256
    &     1.87  &  	0.53  &  	244
    &    9.32	&	9.20	&	364
    &     3.05	&	3.33	&	257
    &      0.68	&	0.57	&	196   \\
    \addlinespace[0.2em]
    \cmidrule(lr){2-17}
    
    & Sink (N=2)
    &  6.24    &   2.94   &603
    &    7.26	 & 2.43 & 	605
    &     25.33	&	22.73	&	818
    &     4.79	&	4.2	&	566
    &    1.33	&	1.38	&	506   \\
    
    & Sink (N=3)
    &  4.31    &  1.95    &363      
    &    6.34 &	2.06	 &354
    &   21.51	&	19.67	&	514
    &     3.73	&	3.49	&	360
    &   0.93	&	0.94	&	298  \\
    
    & Sink (N=4)
    &  3.27&	1.24&	256   
    &    5.50	&   1.64&   	244
    &     19.10	&	17.79	&	364
    &    3.23	&	3.33	&	257
    &    0.69	&	0.65	&	196   \\
    \addlinespace[0.2em]
    \cmidrule(lr){2-17}

    & Unimodal (N=2)
    &   0.65	&0.24	&301
    &     0.89& 	0.3& 	302
    &     7.25	&	6.82	&	409
    &      2.32	&	3.03	&	283
    &    0.21	&	0.45	&	253  \\
    & Unimodal (N=3)
    &   0.92&  	0.4	&  181
    &     1.02& 	0.18	& 177
    &     7.02	&	7.03	&	257
    &     2.06	&	2.87	&	180
    &    0.19	&	0.42	&	149 \\
    & Unimodal (N=4)
    &     0.71	 & 0.37	 & 128
    &    1.07	 &    0.32 &    	122
    &    6.19	&	6.10	&	182
    &     2.11	&	2.78	&	128
    &   0.21	&	0.36	&	98    \\
    \addlinespace[0.2em]
    \cmidrule(lr){2-17}
    
    & Crossmodal (N=2)
    &    5.59	 & 2.96 & 	302
    &     6.55&	2.32	&303
    &     21.30	&	19.93	&	409
    &     4.16	&	3.69	&	283
    &   1.27	&	1.14	&	253 \\
    & Crossmodal (N=3)
    &   3.55&	1.53	&182
    &    5.73	 &  1.85 &  	178
    &    16.81	&	15.78	&	257
    &    3.35	&	3.2	&	180
    &     0.77	&	0.70	&	149  \\
    & Crossmodal (N=4)
    & 2.7	 &1.00	 &128
    &  4.9    & 	1.28    & 	122
    &   14.24	&	13.95	&	182
    &     2.82	&	3.05	&	129
    &     0.76	&	0.64	&	98  \\

    \midrule
    
    \multirow{14}{*}{\makecell{\textbf{Video}\\\textbf{Dominant}}}
    & All
    &    8.21&	13.64&	249
    &   2.43	 &   8.85	 &   250
    &     3.63	&	4.09	&	153
    &    0.46	&	1.86	&	60
    &   -0.05	&	-0.04	&	60    \\
    \addlinespace[0.2em]
    \cmidrule(lr){2-17}

    & Object
    &    4.97	&	8.44	&	149
    &    1.59	&	6.41	&	149
    &    2.07	&	0.40	&	78
    &    0.22	&	1.71	&	7
    &    -0.01	&	-0.06	&	7	 \\
    \addlinespace[0.2em]
    \cmidrule(lr){2-17}
    
    & Random (N=2)
    &  4.15 & 	6.61&	144
    &     1.23	 &5.43	 &148
    &     2.28	&	1.66	&	117
    &     0.24	&	1.81	&	9
    &     -0.02	&	0.02	&	29     \\
    
    & Random (N=3)
    &2.71	&4.28&	87
    &    0.80&  4.20	&  109
    &   1.67	&	-0.72	&	76
    &     0.11	&	1.4	&	4
    &     0	&	0.05	&	17   \\
    
    & Random (N=4)
    &    1.84	 &3.09 &	61
    &     0.44	 &   3.25 &   	85
    &      1.03	&	-2.73	&	52
    &     0.22	&	1.28	&	2
    &    -0.01	&	-0.01	&	13   \\
    \addlinespace[0.2em]
    \cmidrule(lr){2-17}
    
    & Sink (N=2)
    &  5.47	&8.54&	144
    &    2.07&	6.87	&148
    &   3.57	&	3.87	&	117
    &     0.28	&	2.24	&	9
    &    -0.02	&	0	&	29  \\
    
    & Sink (N=3)
    & 4.4 &  	7.12 &  	87
    &     1.62&  	5.89	&  109
    &     3.51	&	3.67	&	76
    &     0.08	&	1.7	&	4
    &   -0.04	&	-0.05	&	17    \\
    
    & Sink (N=4)
    &   3.1	& 6.28& 	61
    &     1.1	& 4.78& 	85
    &     3.29	&	3.29	&	52
    &     0.06	&	1.29	&	2
    &    -0.01	&	0.07	&	13   \\
    \addlinespace[0.2em]
    \cmidrule(lr){2-17}
    
    & Unimodal (N=2)
    &    1.93& 	3.54& 	72
    &     0.35	&   3.43&   	74
    &    -0.01	&	-5.00	&	58
    &    0.18	&	1.31	&	4
    &    0	&	0.03	&	14  \\
    & Unimodal (N=3)
    &     1.72 &	3.19	 &43
    &    0.31& 	3.15& 	54
    &   0.13	&	-4.54	&	38
    &     0.08	&	1.44	&	2
    &    0	&	0.06	&	8 \\
    & Unimodal (N=4)
    &     1.27&	2.8	&30
    &0.24	&2.77	&42
    &  0.18	&	-4.46	&	26
    &  0.08	&	1.33	&	1
    &  -0.02	&	-0.02	&	6   \\
    \addlinespace[0.2em]
    \cmidrule(lr){2-17}
    & Crossmodal (N=2)
    & 3.03	&4.53&	72
    &   1.25 &	4.48	 &74
    &    3.53	&	3.72	&	59
    &     0.26	&	2.19	&	5
    &     -0.02	&	0.01	&	15 \\
    & Crossmodal (N=3)
    &  2.15	& 3.7	& 44
    &   1.01	 &  4.11 &  	55
    &    3.3	&	3.15	&	38
    &     0.2	&	1.6	&	2
    &     0.01	&	0.06	&	9 \\
    & Crossmodal (N=4)
    &1.45	 &   3.02	 &   31
    &   0.63	&3.57	&43
    &   3.04	&	2.86	&	26
    &     0.07	&	1.25	&	1
    &    0.02	&	0.04	&	7    \\

    \bottomrule
    \end{tabular}
}
\end{table*}

%% file: table/ablation_causal_tracing2.tex
\begin{table*}[htbp]
\centering
\tiny
\fontsize{5.5pt}{6.5pt}\selectfont
\caption{Causal tracing results obtained by patching hidden states after the MLP block.
}
\label{tab:mlp_ablation}
\scriptsize
\setlength{\tabcolsep}{3pt}

\newcolumntype{C}[1]{>{\centering\arraybackslash}p{#1}}
\resizebox{\textwidth}{!}
{
    \begin{tabular}{ll|
    C{0.85cm}C{0.85cm}>{\columncolor{gray!15}}C{0.7cm}|
    C{0.85cm}C{0.85cm}>{\columncolor{gray!15}}C{0.7cm}|
    C{0.85cm}C{0.85cm}>{\columncolor{gray!15}}C{0.7cm}|  
    C{0.85cm}C{0.85cm}>{\columncolor{gray!15}}C{0.7cm}|
    C{0.85cm}C{0.85cm}>{\columncolor{gray!15}}C{0.7cm}
    }
    
    \toprule
    \textbf{Modality} & \textbf{Ablation}
    & \multicolumn{3}{c|}{\textbf{Qwen2.5-Omni (7B)}}
    & \multicolumn{3}{c|}{\textbf{Qwen2.5-Omni (3B)}}
    & \multicolumn{3}{c|}{\textbf{video-SALMONN-o1 (7B)}}
    & \multicolumn{3}{c|}{\textbf{video-SALMONN2+ (7B)}}
    & \multicolumn{3}{c}{\textbf{video-SALMONN2+ (3B)}} \\
    
    \cmidrule(lr){3-5}
    \cmidrule(lr){6-8}
    \cmidrule(lr){9-11}   
    \cmidrule(lr){12-14}  
    \cmidrule(lr){15-17}  
    
    &
    & IE$_{\text{clean}}$ $\uparrow$ & IE$_{\text{corr}}$ $\uparrow$ & \#Tokens
    & IE$_{\text{clean}}$ $\uparrow$ & IE$_{\text{corr}}$ $\uparrow$ & \#Tokens
    & IE$_{\text{clean}}$ $\uparrow$ & IE$_{\text{corr}}$ $\uparrow$ & \#Tokens  
    & IE$_{\text{clean}}$ $\uparrow$ & IE$_{\text{corr}}$ $\uparrow$ & \#Tokens
    & IE$_{\text{clean}}$ $\uparrow$ & IE$_{\text{corr}}$ $\uparrow$ & \#Tokens \\
    
    \midrule

    \multirow{14}{*}{\makecell{\textbf{Audio}\\\textbf{Dominant}}}
    & All
    &     4.68	&2.65&1440
    &    3.77	& 1.58	& 1440
    &     5.80	&	9.33	&	1820
    &    2.97	&	3.69	&	1210
    &    0.84	&	1.42	&	1210    \\
    \addlinespace[0.2em]
    \cmidrule(lr){2-17}
    
    & Object
    &    5.05&	2.44&613
    &     2.30	&	-0.03	&	580
    &     2.66	&	5.18	&	852
    &    2.31	&	3.02	&	500
    &    0.45	&	0.93	&	448    \\
    \addlinespace[0.2em]
    \cmidrule(lr){2-17}
    & Random (N=2)
    &   2.92&	1.72&	603
    &   2.52	&	0.94	&	605
    &     2.72	&	4.69	&	818
    &     2.56	&	3.23	&	566
    &    0.51	&	0.73	&	506 \\
    
    & Random (N=3)
    &   2.55	& 1.14	& 363
    &   2.28	&	0.48	&	355
    &     0.93	&	2.44	&	514
    &     2.19	&	3.01	&	360
    &     0.39	&	0.50	&	298   \\
    
    & Random (N=4)
    &   1.89& 	0.77	& 256
    &     1.15	&	0.59	&	244
    &     0.97	&	1.49	&	364
    &     2.09	&	2.91	&	257
    &    0.33	&	0.43	&	196 \\
    \addlinespace[0.2em]
    \cmidrule(lr){2-17}
    & Sink (N=2)
    &  3.62    &  1.81    &603
    &     3.42	&1.35	&605
    &      4.19	&	6.08	&	818
    &     2.61	&	3.01	&	566
    &     0.55	&	0.86	&	506      \\
    
    & Sink (N=3)
    &   2.79   &   1.45   &362
    &  3.19	&	1.22	&	354
    &     2.31	&	3.67	&	514
    &     2.12	&	2.75	&	360
    &     0.49	&	0.67	&	298 \\
    
    & Sink (N=4)
    &   2.01&	0.78	&256
    &    2.91	&	1.1	&	244
    &     2.01	&	3.14	&	364
    &     2.14	&	2.66	&	257
    &     0.31	&	0.44	&	196  \\
    \addlinespace[0.2em]
    \cmidrule(lr){2-17}
    
    & Unimodal (N=2)
    &   0.75   &  0.33    &301
    &     0.9	&	0.18	&	302
    &     1.83	&	2.34	&	409
    &      2.1	&	2.73	&	283
    &      0.21	&	0.42	&	253  \\
    & Unimodal (N=3)
    &   0.65   &   0.42   & 181
    &   0.92	&	0.07	&	177
    &    1.18	&	1.30	&	257
    &     1.89	&	2.63	&	180
    &    0.05	&	0.32	&	149  \\
    & Unimodal (N=4)
    &  0.46	&0.46	&128
    &  0.94	&	0.41	&	122
    &     0.91	&	1.56	&	182
    &     1.89	&	2.64	&	128
    &    0.16	&	0.30	&	98     \\
    \addlinespace[0.2em]
    \cmidrule(lr){2-17}
    & Crossmodal (N=2)
    &   3.89   &  1.87    & 301
    &     3.16	&	1.31	&	303
    &     2.04	&	3.52	&	409
    &     2.24	&	2.84	&	283
    &     0.54	&	0.78	&	253 \\
    & Crossmodal (N=3)
    &   2.26   &  1.20    & 181
    &   2.89	&	0.95	&	178
    &    1.14	&	1.98	&	257
    &     1.88	&	2.69	&	180
    &   0.42	&	0.53	&	149     \\
    & Crossmodal (N=4)
    &     1.96& 	0.9& 	128
    &    2.74	&	0.99	&	122
    &      1.05	&	1.67	&	182
    &     1.84	&	2.58	&	129
    &  0.37	&	0.48	&	98    \\
    
    \midrule
    
    \multirow{14}{*}{\makecell{\textbf{Video}\\\textbf{Dominant}}}
    & All
    & 10.66	&	13.29		&249
    &   4.89	&	8.66	&	250
    &   -0.04	&	3.28	&	153
    &    0.22	&	2.25	&	60
    
    &    0.03	&	0.7	&	60\\
    \addlinespace[0.2em]
    \cmidrule(lr){2-17}
    
    & Object
    &  7.49	&	8.21	&	149
    &   3.45	&	6.14	&	149
    &   -0.07	&	-0.49	&	78
    &   0.14	&	1.47	&	7
    &    0.01	&	0.13	&	7  \\
    \cmidrule(lr){2-17}
    
    & Sink (N=2)
    &     8.31&	8.75&	144
    &     4.23	&	6.28	&	148
    &     0.00	&	3.14	&	117
    &     0.37	&	2.21	&	9
    &    0.04	&	0.35	&	29  \\
    & Sink (N=3)
    &  7.69&	7.51&	87
    &     3.56	&	5.45	&	109
    &     0.00	&	2.80	&	76
    &     0.21	&	1.62	&	4
    &     0.02	&	0.24	&	17   \\
    & Sink (N=4)
    & 6.21 &  	6.3 &  	61
    &   2.77	&	4.63	&	85
    &     0.04	&	2.12	&	52
    &     0.15	&	1.46	&	2
    &   0.02	&	0.18	&	13   \\
    \addlinespace[0.2em]
    \cmidrule(lr){2-17}
    & Random (N=2)
    &   7.93&	7.82	&144
    &    3.00	&	5.06	&	148
    &     -0.06	&	0.62	&	117
    &    0.31	&	1.49	&	9
    &     0.02	&	0.42	&	29    \\
    & Random (N=3)
    &   6.21 & 	5.4	 & 87
    &   2.75	&	4.22	&	109
    &    -0.36	&	-1.94	&	76
    &     0.17	&	1.42	&	4
    &    0.03	&	0.25	&	17  \\
    
    & Random (N=4)
    &  4.62&  	3.98&  	61
    &    2.36	&	3.55	&	85
    &    -0.06	&	-2.81	&	52
    &   0.10	&	1.34	&	2
    &     0.03	&	0.25	&	13  \\
    
    \addlinespace[0.2em]
    \cmidrule(lr){2-17}
    & Unimodal (N=2)
    &   4.21&	4.03&	72
    &   1.34	&	3.28	&	74
    &     0.04	&	-4.87	&	58
    &     0.17	&	1.55	&	4
    & 0.02	&	0.33	&	14\\
    
    & Unimodal (N=3)
    &    4.18 &  	3.83 &  	43
    &     1.02	&	2.83	&	54
    &    0.05	&	-4.70	&	38
    &     0.16	&	1.33	&	2
    &     0.03	&	0.23	&	8     \\
    & Unimodal (N=4)
    &   3.12	 &    2.9 &    	30
    &    0.82	&	2.69	&	42
    &    -0.02	&	2.66	&	59
    &     0.06	&	1.32	&	1
    &      0.02	&	0.27	&	6  \\
    \addlinespace[0.2em]
    \cmidrule(lr){2-17}
    
    & Crossmodal (N=2)
    &  5.86& 	5.16& 	72
    &     3.17	&	4.32	&	74
    &     -0.02	&	2.66	&	59
    &     0.29	&	2.05	&	5
    &    0.03	&	0.21	&	15\\
    & Crossmodal (N=3)
    &   4.83&     	4.3	&     44
    &    2.72	&	3.9	&	55
    &    0.02	&	1.94	&	38
    &     0.16	&	1.54	&	2
    &    0.02	&	0.1	&	9 \\
    & Crossmodal (N=4)
    &     3.89	  &   3.75	  &   31
    &    2.20	&	3.87	&	43
    &     -0.06	&	1.10	&	26
    &     0.14	&	1.39	&	1
    &     0.02	&	0.08	&	7     \\
    
    \bottomrule
    \end{tabular}
}
\end{table*}